\documentclass[11pt]{article}

\usepackage[preprint]{acl}

\makeatletter
\renewcommand\outauthor{%
    \begin{tabular}[t]{c}
    \ifacl@anonymize
        \bfseries Anonymous EMNLP submission
    \else
        \bfseries\@author
    \fi
    \end{tabular}}
\makeatother

\usepackage{times}
\usepackage{latexsym}

\usepackage[T1]{fontenc}
\usepackage[utf8]{inputenc}

\usepackage{microtype}
\usepackage{inconsolata}

\usepackage{graphicx}
\usepackage{booktabs}
\usepackage{tabularx}
\usepackage{array}
\usepackage{multirow}
\usepackage{amsmath}
\usepackage{amssymb}
\usepackage{xspace}
\usepackage{enumitem}


\newcommand{\benchmark}{\textsc{FinBalance}\xspace}

\newcommand{\bsexact}{\textsc{BS}\textsubscript{exact}\xspace}
\newcommand{\bsrecon}{\textsc{BS}\textsubscript{recon}\xspace}

\newcommand{\aggGap}{aggregation gap\xspace}

\newcommand{\incCode}{inconsistency-code match rate\xspace}

\title{FinBalance: A Multi-Document Accounting Reconciliation Benchmark}

\author{%
Sasank Tumpati$^{1,*}$ \quad Devansh Agarwal$^{1,*}$ \quad Ayush Kedia$^{1}$ \quad Arjun Neekhra$^{1}$ \\[1pt]
Murari Mandal$^{2}$ \quad Krishna Garg$^{3}$ \quad Yash Sinha$^{1}$ \quad Suman Gupta$^{1}$ \quad Dhruv Kumar$^{1}$ \\[4pt]
\normalfont\small
$^{1}$BITS Pilani \quad $^{2}$KIIT Bhubaneswar \quad $^{3}$University of Oxford \\[3pt]
\normalfont\small $^{*}$Equal contribution.%
}

\begin{document}
\maketitle

\begin{abstract}
Existing financial-NLP benchmarks mostly evaluate prepared artifacts such as filings, tables, or extracted values. Real accounting begins earlier: source documents must be reconciled into cited journal entries, aggregated into a balance sheet, and checked for contradictions. We introduce \benchmark{}, a multi-document accounting reconciliation benchmark built from source-document bundles across eight industries, three period types, and five difficulty levels. Human-authored business scenarios, accounting policies, tax/FX treatments, document schemas, distractors, and inconsistency templates are composed by a deterministic generator whose ledger produces journal entries, balance sheets, and 23 inconsistency-code labels. On a 710-record evaluation split, six contemporary LLMs reach at most 46\% exact final-balance-sheet accuracy. Four models show a 26--41\,pp gap between \bsexact{}, the model's reported balance sheet, and \bsrecon{}, the balance sheet obtained by replaying its entries through our ledger. Models often recover numerically plausible entries but fail to bind them to supporting documents and aggregate them consistently. Citation-pressure prompting barely changes document-linking errors, while ledger-feedback ablations substantially improve reported balance sheets and expose inconsistency-detection trade-offs. Expert finance reviewers validate the benchmark design and labels.
\end{abstract}

\section{Introduction}
\label{sec:intro}

LLMs are increasingly applied to financial tasks, but the dominant evaluation paradigm targets \emph{comprehension} of already-prepared artifacts. FinQA \citep{chen-etal-2021-finqa}, ConvFinQA \citep{chen-etal-2022-convfinqa}, TAT-QA \citep{zhu-etal-2021-tatqa}, and DocFinQA \citep{reddy-etal-2024-docfinqa} pose numerical questions over excerpted tables and text. FinanceBench \citep{islam2023financebench}, FinanceMATH \citep{zhao-etal-2024-financemath}, FinBen \citep{xie2024finben}, and FinAuditing \citep{wang2026finauditing} test analysis or auditing over prepared statements. In each case the financial statement already exists; the model reasons \emph{over} it.

\begin{table*}[t!]
\centering
\scriptsize
\setlength{\tabcolsep}{2.4pt}
\renewcommand{\arraystretch}{1.06}
\newcommand{\yes}{\(\checkmark\)}
\newcommand{\partly}{\(\triangle\)}
\newcommand{\no}{--}
\begin{tabularx}{\linewidth}{@{}>{\raggedright\arraybackslash}p{2.05cm}>{\raggedright\arraybackslash}p{3.25cm}>{\raggedright\arraybackslash}p{2.65cm}ccccccc@{}}
\toprule
Benchmark & Evidence given to model & Target output & QA & \shortstack{Long/\\multi ctx.} & JE & BS & \shortstack{Doc/span\\grounding} & Incons. & \shortstack{Det.\\replay} \\
\midrule
FinQA \citep{chen-etal-2021-finqa}
& Extracted annual-report table and nearby text
& Numeric answer plus executable reasoning program
& \yes & \partly & \no & \no & \no & \no & \no \\
ConvFinQA \citep{chen-etal-2022-convfinqa}
& Financial report context with conversational turns
& Turn-level numeric answers and programs
& \yes & \partly & \no & \no & \no & \no & \no \\
TAT-QA \citep{zhu-etal-2021-tatqa}
& One hybrid table-text context from annual reports
& Answer with arithmetic/operator supervision
& \yes & \partly & \no & \no & \partly & \no & \no \\
DocFinQA \citep{reddy-etal-2024-docfinqa}
& Full SEC filing attached to FinQA-style questions
& Long-context numeric answer/program
& \yes & \yes & \no & \no & \partly & \no & \no \\
MultiHiertt \citep{zhao-etal-2022-multihiertt}
& Multiple hierarchical tables plus report text
& Numeric answer with supporting facts
& \yes & \yes & \no & \no & \yes & \no & \no \\
FinanceBench \citep{islam2023financebench}
& Public filings and earnings-call evidence
& Open-book financial QA answer
& \yes & \yes & \no & \no & \yes & \no & \no \\
FinanceMath \citep{zhao-etal-2024-financemath}
& Finance-domain word problems with text/table inputs
& Knowledge-intensive math solution
& \partly & \partly & \no & \no & \no & \no & \no \\
FinBen \citep{xie2024finben}
& Suite of finance datasets across IE, QA, RAG, forecasting
& Task-specific outputs across 24 tasks
& \partly & \partly & \no & \no & \partly & \no & \no \\
FinDABench \citep{liu-etal-2025-findabench}
& Financial tables, charts, and documents
& Data-analysis answers over prepared artifacts
& \partly & \partly & \no & \no & \partly & \no & \no \\
FinTagging \citep{wang2025fintagging}
& Financial reports with facts to extract and structure
& Tagged financial facts / structured fields
& \no & \partly & \no & \no & \yes & \no & \no \\
FinRule-Bench \citep{malarkkan2026finrulebench}
& Financial tables plus accounting principles
& Rule-grounded reasoning answer
& \partly & \partly & \no & \no & \partly & \partly & \no \\
FinMaster \citep{jiang2025finmaster}
& Workflow-style finance tasks across roles
& Task-specific financial workflow answers
& \partly & \partly & \partly & \partly & \partly & \partly & \no \\
FinAuditing \citep{wang2026finauditing}
& Real XBRL filings with taxonomy-defined structure
& Semantic, relational, and numerical auditing judgments
& \no & \yes & \no & \no & \yes & \yes & \no \\
\midrule
\benchmark{} (this work)
& Source-document bundle with OCR text and posting/support/distractor docs
& Cited journal entries, final balance sheet, inconsistency code
& \no & \yes & \yes & \yes & \yes & \yes & \yes \\
\bottomrule
\end{tabularx}
\caption{Related-benchmark comparison. \textbf{QA} marks question-answering or problem-solving benchmarks; \textbf{JE} and \textbf{BS} mark construction of journal entries and a final balance sheet; \textbf{Incons.} marks explicit contradiction or consistency detection; \textbf{Det. replay} marks labels and evaluation produced by replaying a deterministic ledger. \yes{} = central objective, \partly{} = adjacent or subset capability, -- = not part of the benchmark objective.}
\label{tab:prior-work-comparison}
\end{table*}

Real accounting workflows begin one step earlier. A bundle of source documents---invoices, bank statements, payment notices, contracts, schedules, exemption certificates---must be reconciled: each transaction extracted as a balanced journal entry citing its supporting documents, the entries aggregated into a balance sheet that respects the accounting equation, and contradictions among documents detected and classified rather than silently resolved. We introduce \benchmark{}, a multi-document accounting reconciliation benchmark built from source-document bundles with OCR text across eight industries, three period types, and five difficulty levels, with concept-coverage flags for revenue recognition under ASC~606, leases, deferred tax, asset disposals, multi-jurisdictional sales tax, foreign-currency remeasurement, and tax exemptions.

\benchmark{} is both a benchmark resource and a diagnostic testbed. The resource itself is nontrivial: it requires explicit modeling of business processes, evidence documents, accounting policy, document dependencies, distractors, and contradiction types before any data can be generated. This makes the benchmark a useful controlled prerequisite for real deployment: if a model cannot reconcile clean, fully specified document bundles with exact ground truth, it is unlikely to be reliable on real accounting work, where layouts, OCR, missing evidence, policy choices, and institutional conventions are messier. We therefore pair the release with our own evaluation suite---six contemporary-model baselines, citation ablations, context-vs.-logic stress tests, passive-tool tests, best-of-3 selection, and a forced ledger-feedback diagnostic---to show what current models can and cannot do on this document-grounded accounting task. In this evaluated panel, the benchmark is already challenging: the highest exact final-balance-sheet score is 46\%, and strict document-grounded end-to-end accuracy remains low.

\benchmark{} is synthetic by construction, but its accounting knowledge is human-authored rather than model-invented. We first spend the design effort where accountants would: defining business scenarios across industries, document schemas, a chart of accounts, accounting policies, tax and FX treatments, support-document dependencies, distractor templates, and contradiction templates. These specifications define the space of valid records and the boundaries of amounts, dates, counterparties, document roles, and scenario combinations. The generator then performs the repetitive composition step: it samples from this human-written accounting specification, fills document fields, selects compatible evidence and distractors, renders source documents with OCR text, and uses the fixed double-entry ledger to compute the expected journal entries and balance sheet. This makes it possible to create many controlled combinations that would be expensive to hand-author one by one, while retaining exact labels: every expected entry, balance sheet, opening trial balance, and inconsistency code is computable, auditable, and reproducible from the specification.

We report two complementary balance-sheet metrics. \bsexact{} measures whether the model's own self-reported balance sheet matches ground truth exactly. \bsrecon{} instead parses the journal entries produced by the model, applies those entries to the opening balance using our deterministic ledger, and checks whether the resulting balance sheet matches ground truth. The two diverge when the model writes entries with the right accounts and amounts, but its self-aggregated balance sheet is wrong. Evaluating six contemporary LLMs (Gemini~3 Flash, GPT-5, Claude Haiku~4.5, Grok-4.3, DeepSeek Chat, Qwen~3 235B) yields three findings. \textbf{(i)}~An \aggGap{}: four of six show $26$--$41$\,pp between \bsrecon{} and \bsexact{}. The remaining two lack the gap for qualitatively different reasons: GPT-5 is comparatively self-consistent, while Grok-4.3 has a smaller gap but lower entry recall (\S\ref{sec:analysis}). \textbf{(ii)}~Document-linking is the dominant entry-level failure: on the anchor ablation split, 52\% of journal entries with correct account and amount cite the wrong documents, a rate that barely moves under explicit citation-pressure prompting ($-1.4$\,pp), implicating retrieval-disambiguation rather than prompt compliance. \textbf{(iii)}~When we use the deterministic ledger as a diagnostic tool and return its balance-sheet deltas to the model, \bsexact{} rises by $+30$ to $+33$\,pp across three model families; the same tool feedback also reveals a trade-off, degrading inconsistency-code detection by $-13$ to $-52$\,pp on smaller models.

\textbf{Contributions.}
\textbf{(1)}~A multi-document accounting reconciliation benchmark with deterministic ground truth, 23 inconsistency codes, and a released generator for fresh stratified samples. \textbf{(2)}~A six-model evaluation on 710 records revealing a self-consistency \aggGap{}, with the highest evaluated model reaching 46\% \bsexact{}, plus failure breakdowns showing which industries, ledger families, concept flags, and error types drive the aggregate scores. \textbf{(3)}~A diagnostic ablation suite covering citation pressure, context stress, passive tools, best-of-3 selection, and forced ledger feedback. \textbf{(4)}~Expert validation by practising finance professionals, including an independently double-reviewed subset and a full-coverage review by a certified accountant spanning every compact scenario at least once.

\section{Related Work}
\label{sec:related}

\paragraph{Financial-NLP question answering.}
FinQA \citep{chen-etal-2021-finqa} and ConvFinQA \citep{chen-etal-2022-convfinqa} pose numerical questions over single financial reports and their tabular content, with the latter requiring conversational multi-turn reasoning. TAT-QA \citep{zhu-etal-2021-tatqa} similarly evaluates models on hybrid tabular-and-text questions over filings. DocFinQA \citep{reddy-etal-2024-docfinqa} extends this paradigm to longer financial documents, and MultiHiertt \citep{zhao-etal-2022-multihiertt} targets reasoning over multiple hierarchical tables and accompanying text. These benchmarks share the assumption that a structured artifact---a table extracted from a filing, a single report---already exists, and the model's task is to answer questions \emph{over} it. \benchmark{} differs structurally: there is no prepared statement, only a bundle of raw source documents with OCR text from which the statement must be constructed.

\paragraph{Financial benchmark suites.}
FinanceBench \citep{islam2023financebench}, FinanceMATH \citep{zhao-etal-2024-financemath}, FinBen \citep{xie2024finben}, FinDABench \citep{liu-etal-2025-findabench}, FinTagging \citep{wang2025fintagging}, FinRule-Bench \citep{malarkkan2026finrulebench}, FinMaster \citep{jiang2025finmaster}, and FinAuditing \citep{wang2026finauditing} broaden financial evaluation to filings, math, tagging, taxonomy-structured auditing, accounting principles, and workflow-style tasks (Table~\ref{tab:prior-work-comparison}). They still evaluate answers, tags, judgments, or prepared-statement reasoning rather than requiring models to construct a cited double-entry ledger and balance sheet from source-document bundles. \benchmark{} fills this gap: its ground truth is a computed ledger, and the consistency between journal entries and their aggregated balance sheet is directly measurable.

\paragraph{Inference-time diagnostics.}
Self-consistency \citep{wang2023selfconsistency}, refinement methods \citep{madaan2023selfrefine,shinn2023reflexion}, trained verifiers \citep{cobbe2021gsm8k,lightman2024verify}, retrieval augmentation \citep{lewis2020rag}, long-context analyses \citep{liu2024lostmiddle}, and tool-use agents \citep{yao2023react} motivate our diagnostics. We do not introduce a general verifier; we use the benchmark ledger, evidence-only visibility, context stress, and tool conditions to localize failures in this accounting task.

\section{Benchmark Design}
\label{sec:benchmark}

\begin{figure*}[t]
\centering
\includegraphics[width=0.96\textwidth]{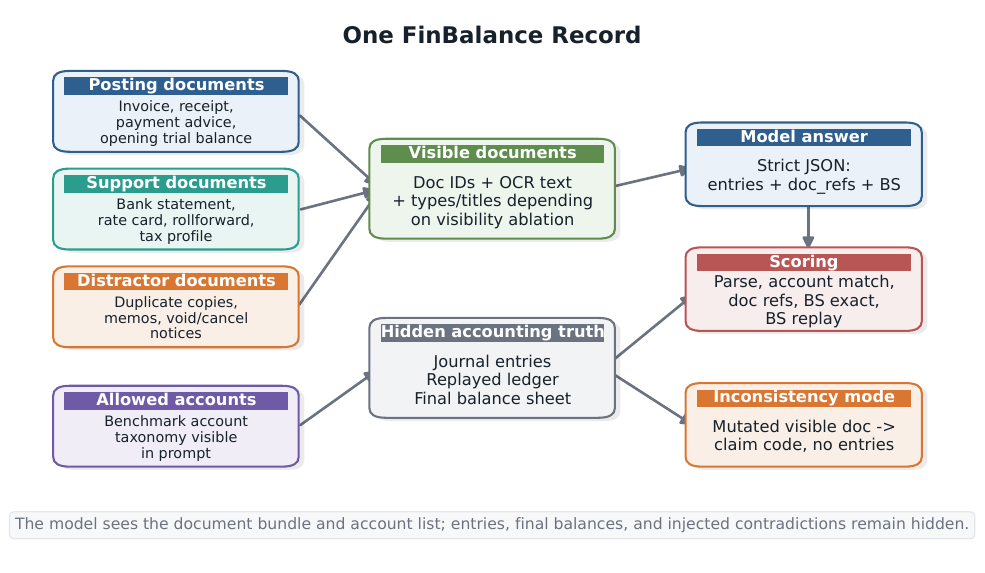}
\caption{One \benchmark{} record. The model sees a visible document bundle of posting, support, adjustment, and distractor documents plus the allowed account taxonomy; hidden journal entries, balance sheets, and inconsistency labels remain available only to the deterministic scorer. For each visible document the baseline prompt contains only its id, type, title, and OCR text (the \texttt{ocr\_only} ablation drops type and title); the \texttt{posting}/\texttt{support}/\texttt{distractor} role tags are used only by the generator and scorer and are never shown to the model.}
\label{fig:dataset-packet}
\end{figure*}

\paragraph{Task.}
A \benchmark{} record is a source-document bundle with OCR text covering one fiscal period for a synthetic business: \emph{posting} documents (invoices, bills, statements, payment notices, vouchers), \emph{support} documents (contracts, schedules, rate sheets, certificates), \emph{adjustment} documents (period-end accruals, rollforwards, reclassifications), an \emph{opening trial balance}, and \emph{distractor} documents that are neutralised but visually plausible. Given the document bundle, the opening trial balance, and the allowed account list, the model outputs JSON with (i)~journal entries (balanced debit/credit postings with account, amount, date, and a \texttt{doc\_refs} list citing supporting documents); (ii)~the final balance sheet structured by section; and (iii)~an inconsistency flag, code, and justification.

\paragraph{Deterministic ledger ground truth.}
Accounting involves policy choices; \benchmark{} makes those choices explicit. Given the fixed chart of accounts, accounting policies, tax assumptions, and transaction generator, labels are deterministic. The human-written layer defines business scenarios, document schemas, tax and FX rules, lease/revenue/deferred-tax treatments, support-document dependencies, distractors, and inconsistency perturbations. The generator selects scenarios, builds subledgers, renders PDF-style documents and OCR text, validates clean records, and replays the ledger to produce entries and the final balance sheet. Appendix Figure~\ref{fig:generation-inference} diagrams this generation-and-scoring workflow; Appendix Figures~\ref{fig:dataset-samples}--\ref{fig:dataset-fx-clean} show complete sample records. Every label is checked against double-entry invariants; no record is hand-labeled after generation.

\paragraph{Coverage.}
Records are stratified over eight industries---professional services, field services, retail, wholesale distribution, healthcare clinics, property management, manufacturing, and subscription SaaS---three period types (month, quarter, year), and five difficulty levels (L1--L5). Difficulty scales document count, cross-document dependencies, jurisdictional complexity, and temporal lookback. \emph{Concept flags} mark the presence of ASC~606 revenue recognition, lease accounting, deferred tax, asset disposals, FX scenarios, multi-jurisdictional tax with exemptions, and rollforwards. The main evaluation split used in this paper contains $710$ records: four standard records for every industry-period-difficulty cell ($480$ total) and ten forced-inconsistency records for every negative-control code ($230$ total). We retain a compact $143$-record split for ablations and fast smoke tests. Finally, the generator is deterministic given a seed and is exposed through user-facing CLI scripts and a Python API, so users can draw fresh stratified samples, create held-out evaluation sets, or extend the scenario space without hand-labeling every record.

\paragraph{Forced-inconsistency records.}
A 23-code taxonomy of deliberate contradictions covers banking, billing, contract, jurisdictional, and reconciliation errors, including closing-balance conflicts, invoice-total mismatches, tax-total mismatches, and invalid exemption certificates. These records extend the benchmark beyond producing a balance sheet: they test whether a model can first recognize that the source documents are internally inconsistent and that a valid balance sheet should not be fabricated, a failure mode that appears routinely in real reconciliation workflows. In a forced-inconsistency record the generator first builds a clean record, then applies a code-specific perturbation that makes the document bundle unreconcilable. The model must set \texttt{has\_inconsistency=True}, emit the correct code, and not fabricate a reconciled balance sheet. Forced-inconsistency records are 32\% of the main evaluation split; Appendix Table~\ref{tab:inconsistency-code-diagnostics} explains every code and reports per-code detection rates.

\paragraph{Metrics.}
For standard records, \bsexact{} measures whether the model's self-reported balance sheet matches ground truth at $0.01$ tolerance. \bsrecon{} parses the model's journal entries, applies those entries to the opening balance with our deterministic ledger, and checks whether the resulting balance sheet matches ground truth; the gap $\bsrecon{}-\bsexact{}$ is the \aggGap{}. Entry matching is posting-level: a strict match requires debit account, credit account, rounded amount, and sorted \texttt{doc\_refs} all correct; the lenient variant ignores \texttt{doc\_refs}. For forced-inconsistency records, balance-sheet and entry metrics are not applicable because the correct output is an inconsistency flag/code with empty entries and balance sheet. We therefore report \incCode{} separately as the rate of correct flag-plus-code over the $230$ negative controls. Parse rate is reported over all records. Appendix~\ref{appx:metrics-detail} gives the full matching procedure.

\paragraph{Ledger-feedback diagnostic.}
\label{sec:verifier}
One ablation repurposes the ledger as inference-time feedback to test whether a model can repair its own reported balance sheet when shown a deterministic consistency check. The harness parses a candidate JSON answer, replays its entries, computes the implied balance sheet, and compares it section-by-section against the model's reported balance sheet. If they disagree, the harness returns per-section deltas, e.g., ``\texttt{current\_assets: submitted=\$48{,}500; replayed=\$51{,}300; delta=\$-2{,}800}''. It gives no ground-truth values or entry-level hints, and the model receives one revision turn.

\section{Experimental Setup}
\label{sec:setup}

\paragraph{Models and decoding.}
We evaluate six contemporary LLMs: Gemini~3 Flash (anchor; full ablation matrix), GPT-5 (low reasoning effort), Claude Haiku~4.5, Grok-4.3, Qwen~3 235B, and DeepSeek Chat. This panel is intended to cover a range of closed, open, general-purpose, and specialist model families. All six receive the full main evaluation split ($n{=}710$) at baseline. The ledger-feedback diagnostic and other targeted ablations are run on the compact $143$-record ablation split, where every industry-period-difficulty cell and every inconsistency code is represented at least once. Decoding uses temperature $0.0$, except best-of-3 selection (temperature $0.7$, three samples).

\paragraph{Ablations and bootstrap.}
We run four ablation families and report all effects against the same-model baseline: \textbf{(1) Prompt} tests guided solving, self-checking, and stronger citation instructions; model behavior is comparatively less sensitive to these prompt changes. \textbf{(2) Visibility} removes the account list, support documents, metadata, or non-evidence documents; these runs show that accounts and support evidence are necessary, and that oracle evidence helps document grounding. \textbf{(3) Context stress} tests whether errors come from longer context or from the harder accounting logic itself by adding distractors or moving relevant evidence later; the results indicate that context hurts, but that the task remains hard even when context is controlled. \textbf{(4) Tool/repair} tests calculators, search, the ledger-feedback tool, and best-of-3 sampling; passive tools are rarely used, while ledger feedback repairs many reported balance sheets but weakens inconsistency detection. Appendix~\ref{appx:ablation-grid} gives the full grid. For uncertainty, we resample record IDs $5000$ times and recompute each baseline--ablation difference on the same resampled records, yielding intervals for \bsexact{}, \bsrecon{}, strict/lenient joint match, \incCode{}, and document-reference mismatch.

\paragraph{Human validation.}
\label{sec:human-validation}
We released 75 expert-validation records spanning all eight industries, all five difficulty levels, and 15 inconsistency codes. Reviewers were practising finance professionals at two major U.S.\ investment banks and worked from self-contained Markdown records. Records 1--25 were independently double-reviewed. Collapsing acceptable judgments (``Yes''/``Mostly'' or ``Accept''/``Accept with minor fixes'') gives $100\%$ agreement on document analogy, entries, \texttt{doc\_refs}, and the overall verdict, and $96\%$ on difficulty calibration. Strict exact-option agreement is lower where reviewers differ on severity rather than acceptability: $64\%$ on analogy, $96\%$ on entry correctness, $100\%$ on entry completeness, $72\%$ on \texttt{doc\_refs} support, $96\%$ on difficulty, and $80\%$ overall. Separately, a certified accountant independently reviewed the full compact coverage split ($143$ records: $120$ standard + $23$ forced-inconsistency). Because this split contains one record for every industry--period--difficulty cell and one for every inconsistency code, that pass checks each compact scenario at least once. The full-coverage review rejected no record; a small number of records received minor tightening notes on document references or wording, while all $23$ forced-inconsistency records were confirmed to contain a real contradiction. Appendix~\ref{appx:human-protocol} gives the protocol, assignment, and marginals.

\section{Results}
\label{sec:results}

\begin{figure*}[t]
\centering
\includegraphics[width=0.92\textwidth]{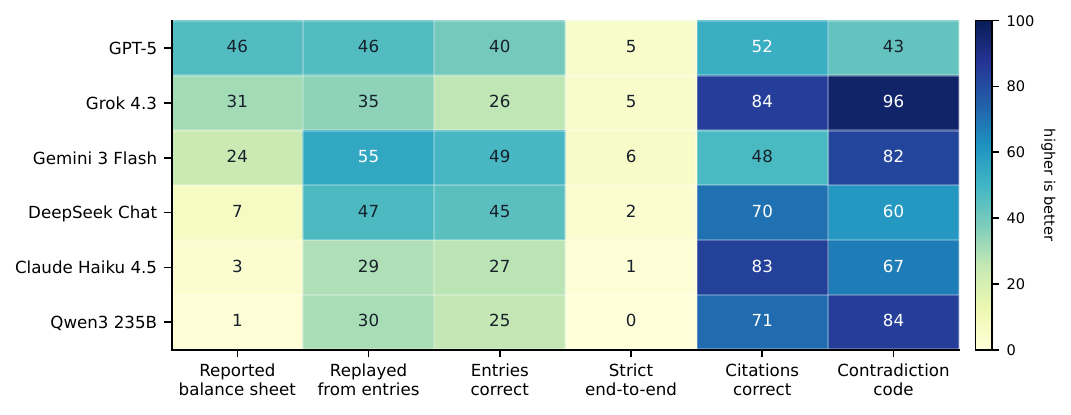}
\caption{Baseline model profile on the 710-record core evaluation split. Balance-sheet and entry metrics use the 480 standard records; \emph{Contradiction code} uses the 230 forced-inconsistency records. \emph{Reported balance sheet} is \bsexact{}; \emph{Replayed from entries} is \bsrecon{}, the balance sheet implied by the model's own journal entries; \emph{Entries correct} ignores document citations; \emph{Strict end-to-end} requires the balance sheet, entries, and citations all to match; \emph{Citations correct} is one minus the document-reference mismatch rate among otherwise correct entries.}
\label{fig:model-accuracy}
\end{figure*}

\benchmark{} is difficult for the models in our evaluation panel. Figure~\ref{fig:model-accuracy} gives the baseline profile: the highest \bsexact{} score is 46\%, strict document-grounded end-to-end accuracy remains low, and the best entry-level accounting scores are much higher than the final balance-sheet scores. The task is therefore not a simple JSON-formatting problem. Models often identify numerically plausible postings, but they lose correctness when those local decisions must be tied back to documents and aggregated into the final statement.

The central pattern in Figure~\ref{fig:model-accuracy} is the gap between \bsexact{} and \bsrecon{}, where \bsrecon{} parses the model's journal entries, applies them to the opening balance with our ledger, and checks the resulting balance sheet against ground truth. In four of six models the two metrics disagree dramatically: DeepSeek Chat reaches $0.473$ \bsrecon{} but only $0.067$ \bsexact{}, a $40.6$\,pp \aggGap{}. Gemini~3 Flash ($+31.3$\,pp), Qwen~3 235B ($+29.0$\,pp), and Claude Haiku~4.5 ($+26.3$\,pp) show the same pattern. Since \bsrecon{} ignores document support, the claim is not that entries are audit-grounded; it is that models often recover plausible entries, fail to bind them to evidence, and then fail to aggregate them into a self-consistent reported statement. GPT-5 is comparatively self-consistent ($46.5\%$ \bsexact{}, $46.3\%$ \bsrecon{}), while Grok-4.3 has a smaller gap but lower entry recall (\S\ref{sec:analysis}).

\begin{figure*}[t]
\centering
\includegraphics[width=0.98\textwidth]{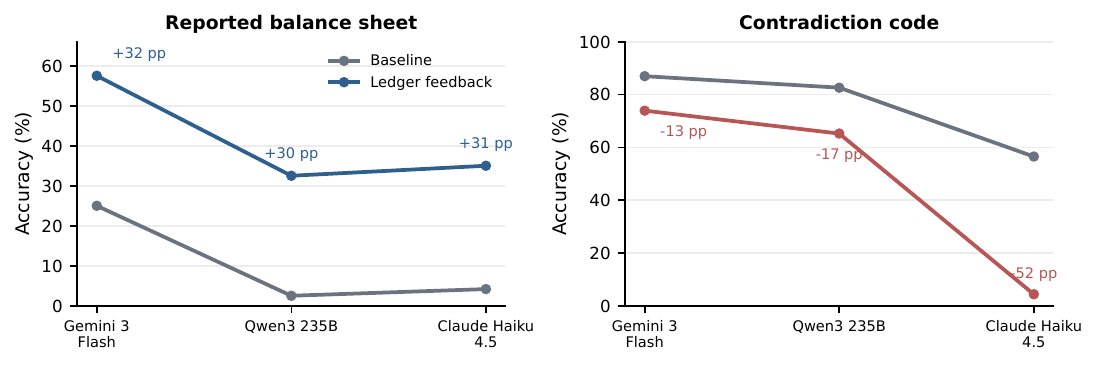}
\caption{One-pass ledger-feedback effect on the compact ablation split. Lines compare each model's baseline with the same model after ledger feedback; labels are percentage-point changes. The tool strongly improves reported balance-sheet accuracy (left) but can reduce contradiction-code accuracy (right). GPT-5 and Grok-4.3 are omitted because they do not exhibit the same aggregation-gap mechanism (Table~\ref{tab:no-gap-mechanism}).}
\label{fig:verifier-effects}
\end{figure*}

If the gap is partly a self-consistency failure rather than only an entries deficit, then applying the model's own entries through the ledger should expose it. We test this with ledger-feedback (\S\ref{sec:verifier}): the ledger compares the model's reported balance sheet against the one its entries imply, returns per-section deltas, and gives the model one revision turn. Figure~\ref{fig:verifier-effects} shows that \bsexact{} gains $+30$ to $+33$\,pp across three gap-having model families, all significant at $95\%$ paired bootstrap. The feedback gives no ground-truth values or entry-level hints---only the model's own implied-versus-reported deltas.

The same diagnostic also reveals a real cost. Figure~\ref{fig:verifier-effects} (right) shows the feedback loop's effect on \incCode{} on the compact ablation split. Gemini~3 Flash loses $-13$\,pp (not significant), while Qwen~3 235B loses $-17$\,pp and Claude Haiku~4.5 loses $-52$\,pp. When a forced-inconsistency record's first-pass submission does not raise the flag, ledger feedback sends balance-sheet deltas instead of giving the model a chance to reconsider whether the documents themselves contradict; the deltas bias the next turn toward reconciling rather than flagging. The trade-off is mild for strong detectors and severe for weak ones.

\begin{figure}[t]
\centering
\includegraphics[width=\linewidth]{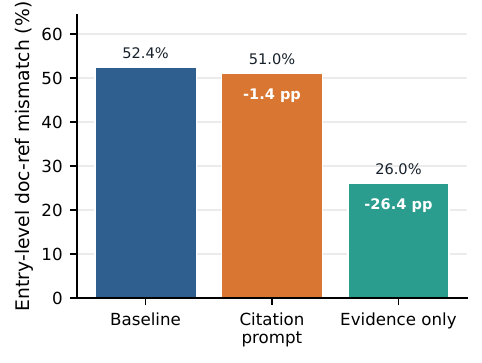}
\caption{Document-linking failure persists under explicit citation pressure, but drops under an evidence-only oracle.}
\label{fig:doc-refs-persistence}
\end{figure}

The aggregation gap is one of two large failure modes; the other is document-linking. On the compact anchor split, $52\%$ of journal entries with correct account and amount cite the \emph{wrong} supporting documents, collapsing strict joint accuracy to $5\%$ despite $57\%$ entry-match when citations are ignored. Citation-pressure prompting changes the mismatch rate by only $-1.4$\,pp (CI $[-2.9, 0.0]$), while an \emph{evidence-only} oracle lowers it by $-26.4$\,pp (CI $[-29.8, -23.0]$). The failure is retrieval-disambiguation, not prompt compliance (Figure~\ref{fig:doc-refs-persistence}).

An alternative is to sample multiple candidates and pick. On the 45 standard records where Gemini's entries replay to the correct balance sheet but its reported balance sheet is wrong, a ledger-aware best-of-3 selector raises \bsexact{} from $0\%$ to $28.9\%$ (CI $[15.6,42.2]$), while one ledger-feedback revision reaches $84.4\%$ (CI $[73.3,93.3]$). Picking a better draft recovers part of the gap; revising after structured ledger feedback recovers much more (Appendix Figure~\ref{fig:gap-repair}).

Table~\ref{tab:ablation-summary} distills the ablation results by diagnostic question. Forced ledger-feedback variants are the only diagnostics that move \bsexact{} substantially upward; removing scaffolding (allowed accounts, support docs) is the most damaging ablation. Prompt-only variants, including a stronger citation instruction and a guided solve protocol, do not close the large effects above. The full anchor-model delta plot is in Appendix Figure~\ref{fig:ablation-deltas}.

\begin{table}[t]
\centering
\scriptsize
\setlength{\tabcolsep}{2.4pt}
\renewcommand{\arraystretch}{1.08}
\begin{tabularx}{\linewidth}{@{}p{1.7cm}p{2.55cm}X@{}}
\toprule
Question & Main effect & Interpretation \\
\midrule
Prompt & Guided/self-check prompts change \bsexact{} by at most $2.5$\,pp; citation pressure changes ref mismatch by $-1.4$\,pp. & Failures are not primarily prompt-compliance artifacts. \\
Evidence & Removing accounts drops \bsrecon{} by $47.5$\,pp; removing support docs drops it by $38.3$\,pp. & Account taxonomy and support evidence are essential. \\
Grounding & Evidence-only visibility lowers ref mismatch by $26.4$\,pp. & Citation failure is mostly evidence selection, not wording. \\
Context & Adding distractors to evidence-only visibility costs $16$--$18$\,pp \bsrecon{} and saturates. & Longer context hurts, but does not explain the task difficulty. \\
Repair & One ledger-feedback turn raises \bsexact{} by $32.5$\,pp on Gemini; best-of-3 raises it by $28.9$\,pp on the targeted subset. & Global ledger checks repair many reported balance sheets. \\
\bottomrule
\end{tabularx}
\caption{Compact ablation summary. Deltas are percentage points against the Gemini~3 Flash anchor baseline unless a targeted subset is stated.}
\label{tab:ablation-summary}
\end{table}

\begin{figure}[t]
\centering
\includegraphics[width=\linewidth]{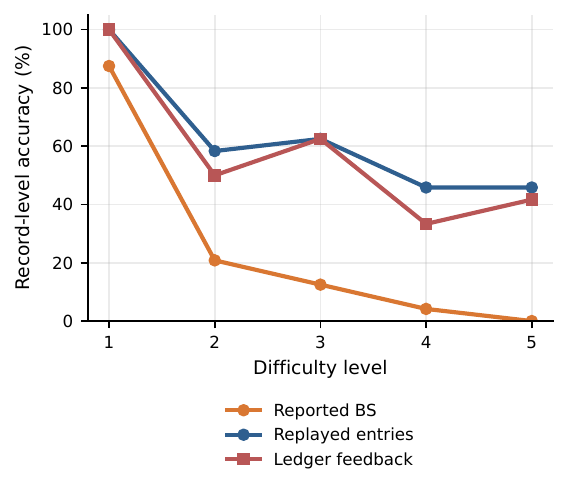}
\caption{Accuracy by difficulty level on the anchor model. As records move from L1 to L5, \bsexact{} collapses much more sharply than the context-stress penalty in Figure~\ref{fig:context-stress}; \bsrecon{} remains higher, showing that entry replay remains easier than reporting the final balance sheet.}
\label{fig:difficulty-trend}
\end{figure}

\begin{figure}[t]
\centering
\includegraphics[width=\linewidth]{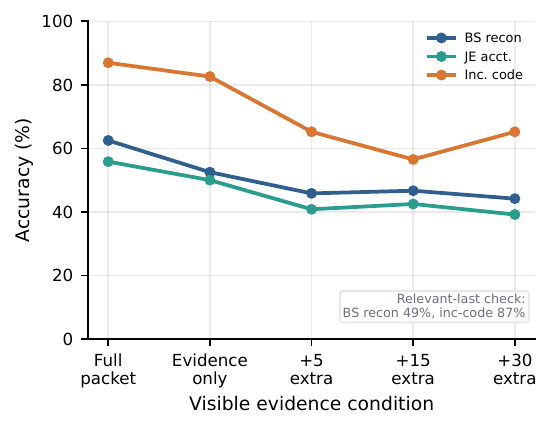}
\caption{Context stress on the anchor model, designed to separate longer-context effects from the underlying accounting-logic difficulty. The task remains hard under controlled evidence; distractors hurt most when first introduced and then saturate.}
\label{fig:context-stress}
\end{figure}

Figures~\ref{fig:difficulty-trend} and~\ref{fig:context-stress} separate two concerns that are easy to conflate: is the benchmark hard because the prompt contains too much text, or because the model must apply the right accounting logic to the right evidence? Difficulty produces the larger effect: baseline \bsexact{} falls from $88\%$ at L1 to $0\%$ at L5, while \bsrecon{} also declines from $100\%$ to $46\%$. By contrast, adding $5$, $15$, or $30$ distractors to an evidence-only document bundle decreases \bsrecon{} by $16.7$, $15.8$, and $18.3$\,pp respectively, with the penalty appearing quickly and then saturating. Thus context contributes, but it is not the main explanation; accounting dependencies and document choice remain binding even when context is controlled.

\section{Analysis}
\label{sec:analysis}

\paragraph{Why two models lack the gap.}
The no-gap models differ. \mbox{GPT-5 submits} $91\%$ of expected entries and has \bsexact{} essentially equal to \bsrecon{} ($0.465$ vs.\ $0.463$). \mbox{Grok-4.3 submits} only $40\%$ of expected entries, so its balance sheet is internally aligned but incomplete (Appendix Table~\ref{tab:no-gap-mechanism}).

\paragraph{Where the gap concentrates.}
On the anchor model, \bsexact{} drops from $0.95$ at L1 to $0.02$ at L5 while \bsrecon{} declines more gently; the \aggGap{} peaks near L3 at roughly $46$\,pp (Figure~\ref{fig:difficulty-trend}). Ledger feedback narrows the gap at every level. Industry, ledger-family, and concept slices in Appendix~\ref{appx:per-concept} and Figure~\ref{fig:failure-slices} show where the benchmark stresses models beyond the aggregate score.

\paragraph{Error taxonomy.}
On the 480 standard records, document-linking failures dominate the anchor model: $93.1\%$ of records and $3890$ entries, compared with $1033$ omissions, $584$ extras, $202$ arithmetic errors, and $85$ account-selection errors. The hard part is grounding plausible entries to the right evidence. Appendix Figure~\ref{fig:failure-slices} breaks failures down by industry, ledger family, and error type.

\section{Discussion}
\label{sec:discussion}

Source-document accounting is not solved by long-context retrieval. Distractors hurt, but controlled-evidence records still degrade as cross-document accounting dependencies grow: models must select evidence, form balanced entries, and aggregate them. Repair ablations target the second stage. Ledger feedback helps more than best-of-3, so the issue is not just insufficient sampling; however, strict document-grounded accuracy stays low, and ledger-consistency feedback can reduce inconsistency detection by pushing models to reconcile contradictory documents.

\section{Conclusion}
\label{sec:conclusion}

\benchmark{} makes document-grounded accounting reconciliation a concrete, computable benchmark: models must select evidence, form cited ledger entries, and aggregate a balance sheet. Across six LLMs, this full pipeline remains hard; replayed entries often outscore reported balance sheets, while strict citation-grounded accuracy stays low. The released generator supports new stratified samples and larger splits from the same human-authored accounting specification.

\section*{Limitations}

\paragraph{Synthetic construction.}
\benchmark{} is synthetic in the way a good controlled benchmark should be: it composes human-authored business scenarios, document schemas, accounting policies, tax and FX assumptions, support-document dependencies, distractors, and contradiction templates, then uses code for the repetitive sampling, rendering, and ledger replay. This gives us inspectable labels and controlled counterfactuals. The limitation is document fidelity: real accounting document bundles contain messier layouts, OCR errors, missing pages, handwriting, ambiguous vendor formats, inconsistent naming, and institution-specific artifacts that our rendered documents only approximate. The benchmark therefore measures reasoning over clean OCR text, not end-to-end document understanding in production systems.

\paragraph{Accounting coverage.}
Accounting is broad and policy-dependent. Our labels are deterministic after fixing the chart of accounts, accounting policy, tax assumptions, and transaction generator. We cover a substantial slice of reconciliation work---multi-document postings, period-end adjustments, tax, foreign exchange, leases, deferred tax, revenue schedules, subledgers, distractors, and forced contradictions---but not the full space of GAAP, IFRS, local tax regimes, materiality judgments, industry-specific policy choices, or audit procedures. The benchmark is intended to open a precise document-grounded accounting setting, not to replace a complete accounting standards corpus.

\paragraph{Analysis depth.}
The ablations reveal several robust patterns, including the aggregation gap, weak gains from prompt-only changes, document-linking failures, context/evidence effects, and the ledger-feedback trade-off with inconsistency detection. We analyse these effects at the behavioral level, but we do not establish the internal mechanisms that cause them, nor whether the same failure modes transfer unchanged to other high-precision domains such as legal reconciliation, insurance claims, or scientific lab records. The released harness is intended to support deeper follow-up studies: larger ablation matrices, controlled perturbations, model-family comparisons, and mechanistic or representation-level analyses of why models fail after the relevant evidence is present.

\paragraph{Evaluation scope.}
Headline numbers use a single temperature-$0$ decoding run except for the best-of-3 condition, so the paper emphasizes large paired effects rather than single-point differences. Six models are evaluated on the full 710-record core split; targeted ablations are run on the compact 143-record split, which preserves coverage over every industry--period--difficulty cell and inconsistency code. The aggregation gap is therefore a consistent pattern across this model panel, not a claim about every language model.

\paragraph{Benchmark longevity.}
The public core split may become contaminated as models train on released benchmarks, a risk shared by most static evaluation sets. \benchmark{} mitigates this by releasing the generator, not only a fixed test file: users can sample fresh seeded document bundles from the same human-authored scenario space, create larger train/test splits, and add new industries, document schemas, policies, contradiction templates, and ablation surfaces without hand-labeling every record. We version the released split for comparability; long-term validity comes from periodically generated held-out document bundles and community extensions of the scenario library.

\paragraph{Ethics and AI assistance.}
The benchmark uses synthetic data; no real records, PII, or production accounting data are used. The ARR Responsible NLP checklist is completed in the submission materials; Appendix~\ref{appx:reproducibility} reports reproducibility and release details only. We used generative-AI assistance for coding support and for grammar, wording, and formatting edits while drafting. The benchmark idea, accounting logic, experimental design, result interpretation, paper argument, and final claims were authored and controlled by humans. We did not use AI assistance to create hidden reviewer-facing prompts or instructions intended to manipulate reviewers.

\bibliography{references}

\begin{thebibliography}{21}
\providecommand{\natexlab}[1]{#1}

\bibitem[{Chen et~al.(2021)Chen, Chen, Smiley, Shah, Borova, Langdon, Moussa,
  Beane, Huang, Routledge, and Wang}]{chen-etal-2021-finqa}
Zhiyu Chen, Wenhu Chen, Charese Smiley, Sameena Shah, Iana Borova, Dylan
  Langdon, Reema Moussa, Matt Beane, Ting-Hao Huang, Bryan Routledge, and
  William~Yang Wang. 2021.
\newblock \href {https://doi.org/10.18653/v1/2021.emnlp-main.300} {{F}in{QA}: A
  dataset of numerical reasoning over financial data}.
\newblock In \emph{Proceedings of the 2021 Conference on Empirical Methods in
  Natural Language Processing}, pages 3697--3711. Association for Computational
  Linguistics.

\bibitem[{Chen et~al.(2022)Chen, Li, Smiley, Ma, Shah, and
  Wang}]{chen-etal-2022-convfinqa}
Zhiyu Chen, Shiyang Li, Charese Smiley, Zhiqiang Ma, Sameena Shah, and
  William~Yang Wang. 2022.
\newblock \href {https://doi.org/10.18653/v1/2022.emnlp-main.421}
  {{C}onv{F}in{QA}: Exploring the chain of numerical reasoning in
  conversational finance question answering}.
\newblock In \emph{Proceedings of the 2022 Conference on Empirical Methods in
  Natural Language Processing}, pages 6279--6292. Association for Computational
  Linguistics.

\bibitem[{Cobbe et~al.(2021)Cobbe, Kosaraju, Bavarian, Chen, Jun, Kaiser,
  Plappert, Tworek, Hilton, Nakano, Hesse, and Schulman}]{cobbe2021gsm8k}
Karl Cobbe, Vineet Kosaraju, Mohammad Bavarian, Mark Chen, Heewoo Jun, Lukasz
  Kaiser, Matthias Plappert, Jerry Tworek, Jacob Hilton, Reiichiro Nakano,
  Christopher Hesse, and John Schulman. 2021.
\newblock \href {https://arxiv.org/abs/2110.14168} {Training verifiers to solve
  math word problems}.
\newblock \emph{Preprint}, arXiv:2110.14168.

\bibitem[{Islam et~al.(2023)Islam, Kannappan, Kiela, Qian, Scherrer, and
  Vidgen}]{islam2023financebench}
Pranab Islam, Anand Kannappan, Douwe Kiela, Rebecca Qian, Nino Scherrer, and
  Bertie Vidgen. 2023.
\newblock \href {https://arxiv.org/abs/2311.11944} {{F}inance{B}ench: A new
  benchmark for financial question answering}.
\newblock \emph{Preprint}, arXiv:2311.11944.

\bibitem[{Jiang et~al.(2025)Jiang, Yang, Cui, Jin, Wang, Li, Huang, Sun, and
  Wang}]{jiang2025finmaster}
Junzhe Jiang, Chang Yang, Aixin Cui, Sihan Jin, Ruiyu Wang, Bo~Li, Xiao Huang,
  Dongning Sun, and Xinrun Wang. 2025.
\newblock \href {https://arxiv.org/abs/2505.13533} {{F}in{M}aster: A holistic
  benchmark for mastering full-pipeline financial workflows with {LLM}s}.
\newblock \emph{Preprint}, arXiv:2505.13533.

\bibitem[{Lewis et~al.(2020)Lewis, Perez, Piktus, Petroni, Karpukhin, Goyal,
  K{\"u}ttler, Lewis, Yih, Rockt{\"a}schel, Riedel, and Kiela}]{lewis2020rag}
Patrick Lewis, Ethan Perez, Aleksandra Piktus, Fabio Petroni, Vladimir
  Karpukhin, Naman Goyal, Heinrich K{\"u}ttler, Mike Lewis, Wen-tau Yih, Tim
  Rockt{\"a}schel, Sebastian Riedel, and Douwe Kiela. 2020.
\newblock \href
  {https://proceedings.neurips.cc/paper/2020/hash/6b493230205f780e1bc26945df7481e5-Abstract.html}
  {Retrieval-augmented generation for knowledge-intensive {NLP} tasks}.
\newblock In \emph{Advances in Neural Information Processing Systems 33
  (NeurIPS)}.

\bibitem[{Lightman et~al.(2024)Lightman, Kosaraju, Burda, Edwards, Baker, Lee,
  Leike, Schulman, Sutskever, and Cobbe}]{lightman2024verify}
Hunter Lightman, Vineet Kosaraju, Yura Burda, Harri Edwards, Bowen Baker, Teddy
  Lee, Jan Leike, John Schulman, Ilya Sutskever, and Karl Cobbe. 2024.
\newblock \href {https://openreview.net/forum?id=v8L0pN6EOi} {Let's verify step
  by step}.
\newblock In \emph{The Twelfth International Conference on Learning
  Representations (ICLR)}.

\bibitem[{Liu et~al.(2024)Liu, Lin, Hewitt, Paranjape, Bevilacqua, Petroni, and
  Liang}]{liu2024lostmiddle}
Nelson~F. Liu, Kevin Lin, John Hewitt, Ashwin Paranjape, Michele Bevilacqua,
  Fabio Petroni, and Percy Liang. 2024.
\newblock \href {https://doi.org/10.1162/tacl_a_00638} {Lost in the middle: How
  language models use long contexts}.
\newblock \emph{Transactions of the Association for Computational Linguistics},
  12:157--173.

\bibitem[{Liu et~al.(2025)Liu, Zhao, Jia, Zhuang, Long, Zhou, Zhou, Lan, and
  Chong}]{liu-etal-2025-findabench}
Shu Liu, Shangqing Zhao, Chenghao Jia, Xinlin Zhuang, Zhaoguang Long, Jie Zhou,
  Aimin Zhou, Man Lan, and Yang Chong. 2025.
\newblock \href {https://aclanthology.org/2025.coling-main.48/}
  {{F}in{DAB}ench: Benchmarking financial data analysis ability of large
  language models}.
\newblock In \emph{Proceedings of the 31st International Conference on
  Computational Linguistics}, pages 710--725. Association for Computational
  Linguistics.

\bibitem[{Madaan et~al.(2023)Madaan, Tandon, Gupta, Hallinan, Gao, Wiegreffe,
  Alon, Dziri, Prabhumoye, Yang, Gupta, Majumder, Hermann, Welleck,
  Yazdanbakhsh, and Clark}]{madaan2023selfrefine}
Aman Madaan, Niket Tandon, Prakhar Gupta, Skyler Hallinan, Luyu Gao, Sarah
  Wiegreffe, Uri Alon, Nouha Dziri, Shrimai Prabhumoye, Yiming Yang, Shashank
  Gupta, Bodhisattwa~Prasad Majumder, Katherine Hermann, Sean Welleck, Amir
  Yazdanbakhsh, and Peter Clark. 2023.
\newblock \href
  {https://proceedings.neurips.cc/paper_files/paper/2023/hash/91edff07232fb1b55a505a9e9f6c0ff3-Abstract-Conference.html}
  {Self-refine: Iterative refinement with self-feedback}.
\newblock In \emph{Advances in Neural Information Processing Systems 36
  (NeurIPS)}.

\bibitem[{Malarkkan et~al.(2026)Malarkkan, Choudhury, Zhang, Gupta, Wang, Fu,
  and Zhang}]{malarkkan2026finrulebench}
Arun~Vignesh Malarkkan, Manan~Roy Choudhury, Guangwei Zhang, Vivek Gupta,
  Qingyun Wang, Yanjie Fu, and Denghui Zhang. 2026.
\newblock \href {https://arxiv.org/abs/2603.11339} {{F}in{R}ule-{B}ench: A
  benchmark for joint reasoning over financial tables and principles}.
\newblock \emph{Preprint}, arXiv:2603.11339.

\bibitem[{Reddy et~al.(2024)Reddy, Koncel-Kedziorski, Lai, Krumdick, Lovering,
  and Tanner}]{reddy-etal-2024-docfinqa}
Varshini Reddy, Rik Koncel-Kedziorski, Viet~Dac Lai, Michael Krumdick, Charles
  Lovering, and Chris Tanner. 2024.
\newblock \href {https://doi.org/10.18653/v1/2024.acl-short.42}
  {{D}oc{F}in{QA}: A long-context financial reasoning dataset}.
\newblock In \emph{Proceedings of the 62nd Annual Meeting of the Association
  for Computational Linguistics (Volume 2: Short Papers)}, pages 445--458.
  Association for Computational Linguistics.

\bibitem[{Shinn et~al.(2023)Shinn, Cassano, Berman, Gopinath, Narasimhan, and
  Yao}]{shinn2023reflexion}
Noah Shinn, Federico Cassano, Edward Berman, Ashwin Gopinath, Karthik
  Narasimhan, and Shunyu Yao. 2023.
\newblock \href {https://openreview.net/forum?id=vAElhFcKW6} {Reflexion:
  Language agents with verbal reinforcement learning}.
\newblock In \emph{Advances in Neural Information Processing Systems 36
  (NeurIPS)}.

\bibitem[{Wang et~al.(2023)Wang, Wei, Schuurmans, Le, Chi, Narang, Chowdhery,
  and Zhou}]{wang2023selfconsistency}
Xuezhi Wang, Jason Wei, Dale Schuurmans, Quoc Le, Ed~H. Chi, Sharan Narang,
  Aakanksha Chowdhery, and Denny Zhou. 2023.
\newblock \href {https://openreview.net/forum?id=1PL1NIMMrw} {Self-consistency
  improves chain of thought reasoning in language models}.
\newblock In \emph{The Eleventh International Conference on Learning
  Representations (ICLR)}.

\bibitem[{Wang et~al.(2025)Wang, Ren, Qian, Peng, Wang, Han, Feng, Liu, Huang,
  and Xie}]{wang2025fintagging}
Yan Wang, Yang Ren, Lingfei Qian, Xueqing Peng, Keyi Wang, Yi~Han, Dongji Feng,
  Xiao-Yang Liu, Jimin Huang, and Qianqian Xie. 2025.
\newblock \href {https://arxiv.org/abs/2505.20650} {{F}in{T}agging: An
  {LLM}-ready benchmark for extracting and structuring financial information}.
\newblock \emph{Preprint}, arXiv:2505.20650.

\bibitem[{Wang et~al.(2026)Wang, Wang, Yang, Patel, Zhao, Mo, Peng, Qian, Chen,
  Guti{\'e}rrez-Basulto, Huang, Xiong, Liu, Liu, and Nie}]{wang2026finauditing}
Yan Wang, Keyi Wang, Shanshan Yang, Jaisal Patel, Jeff Zhao, Fengran Mo,
  Xueqing Peng, Lingfei Qian, Yankai Chen, V{\'i}ctor Guti{\'e}rrez-Basulto,
  Jimin Huang, Guojun Xiong, Xiao-Yang Liu, Xue Liu, and Jian-Yun Nie. 2026.
\newblock \href {https://arxiv.org/abs/2510.08886} {{F}in{A}uditing: A
  financial taxonomy-structured multi-document benchmark for evaluating llms}.
\newblock \emph{Preprint}, arXiv:2510.08886.
\newblock To appear, SIGIR 2026 Resource Track.

\bibitem[{Xie et~al.(2024)Xie, Han, Chen, Xiang, Zhang, He, Xiao, Li, Dai,
  Feng, Xu, Kang, Kuang, Yuan, Yang, Luo, Zhang, Liu, Xiong, Deng, Jiang, Yao,
  Li, Yu, Hu, Huang, Liu, Lopez-Lira, Wang, Lai, Wang, Peng, Ananiadou, and
  Huang}]{xie2024finben}
Qianqian Xie, Weiguang Han, Zhengyu Chen, Ruoyu Xiang, Xiao Zhang, Yueru He,
  Mengxi Xiao, Dong Li, Yongfu Dai, Duanyu Feng, Yijing Xu, Haoqiang Kang,
  Ziyan Kuang, Chenhan Yuan, Kailai Yang, Zheheng Luo, Tianlin Zhang, Zhiwei
  Liu, Guojun Xiong, and 15 others. 2024.
\newblock \href {https://openreview.net/forum?id=loDHZstVP6} {{F}in{B}en: A
  holistic financial benchmark for large language models}.
\newblock In \emph{Advances in Neural Information Processing Systems 37
  (Datasets and Benchmarks Track)}.

\bibitem[{Yao et~al.(2023)Yao, Zhao, Yu, Du, Shafran, Narasimhan, and
  Cao}]{yao2023react}
Shunyu Yao, Jeffrey Zhao, Dian Yu, Nan Du, Izhak Shafran, Karthik Narasimhan,
  and Yuan Cao. 2023.
\newblock \href {https://openreview.net/forum?id=WE_vluYUL-X} {{R}e{A}ct:
  Synergizing reasoning and acting in language models}.
\newblock In \emph{The Eleventh International Conference on Learning
  Representations (ICLR)}.

\bibitem[{Zhao et~al.(2022)Zhao, Li, Li, and
  Zhang}]{zhao-etal-2022-multihiertt}
Yilun Zhao, Yunxiang Li, Chenying Li, and Rui Zhang. 2022.
\newblock \href {https://doi.org/10.18653/v1/2022.acl-long.454}
  {{M}ulti{H}iertt: Numerical reasoning over multi hierarchical tabular and
  textual data}.
\newblock In \emph{Proceedings of the 60th Annual Meeting of the Association
  for Computational Linguistics (Volume 1: Long Papers)}, pages 6588--6600.
  Association for Computational Linguistics.

\bibitem[{Zhao et~al.(2024)Zhao, Liu, Long, Zhang, Zhao, and
  Cohan}]{zhao-etal-2024-financemath}
Yilun Zhao, Hongjun Liu, Yitao Long, Rui Zhang, Chen Zhao, and Arman Cohan.
  2024.
\newblock \href {https://doi.org/10.18653/v1/2024.acl-long.693}
  {{F}inance{MATH}: Knowledge-intensive math reasoning in finance domains}.
\newblock In \emph{Proceedings of the 62nd Annual Meeting of the Association
  for Computational Linguistics (Volume 1: Long Papers)}, pages 12841--12858.
  Association for Computational Linguistics.

\bibitem[{Zhu et~al.(2021)Zhu, Lei, Huang, Wang, Zhang, Lv, Feng, and
  Chua}]{zhu-etal-2021-tatqa}
Fengbin Zhu, Wenqiang Lei, Youcheng Huang, Chao Wang, Shuo Zhang, Jiancheng Lv,
  Fuli Feng, and Tat-Seng Chua. 2021.
\newblock \href {https://doi.org/10.18653/v1/2021.acl-long.254} {{TAT}-{QA}: A
  question answering benchmark on a hybrid of tabular and textual content in
  finance}.
\newblock In \emph{Proceedings of the 59th Annual Meeting of the Association
  for Computational Linguistics and the 11th International Joint Conference on
  Natural Language Processing (Volume 1: Long Papers)}, pages 3277--3287.
  Association for Computational Linguistics.

\end{thebibliography}

\appendix

\section{Dataset Examples}
\label{appx:dataset-examples}

\begin{figure*}[t]
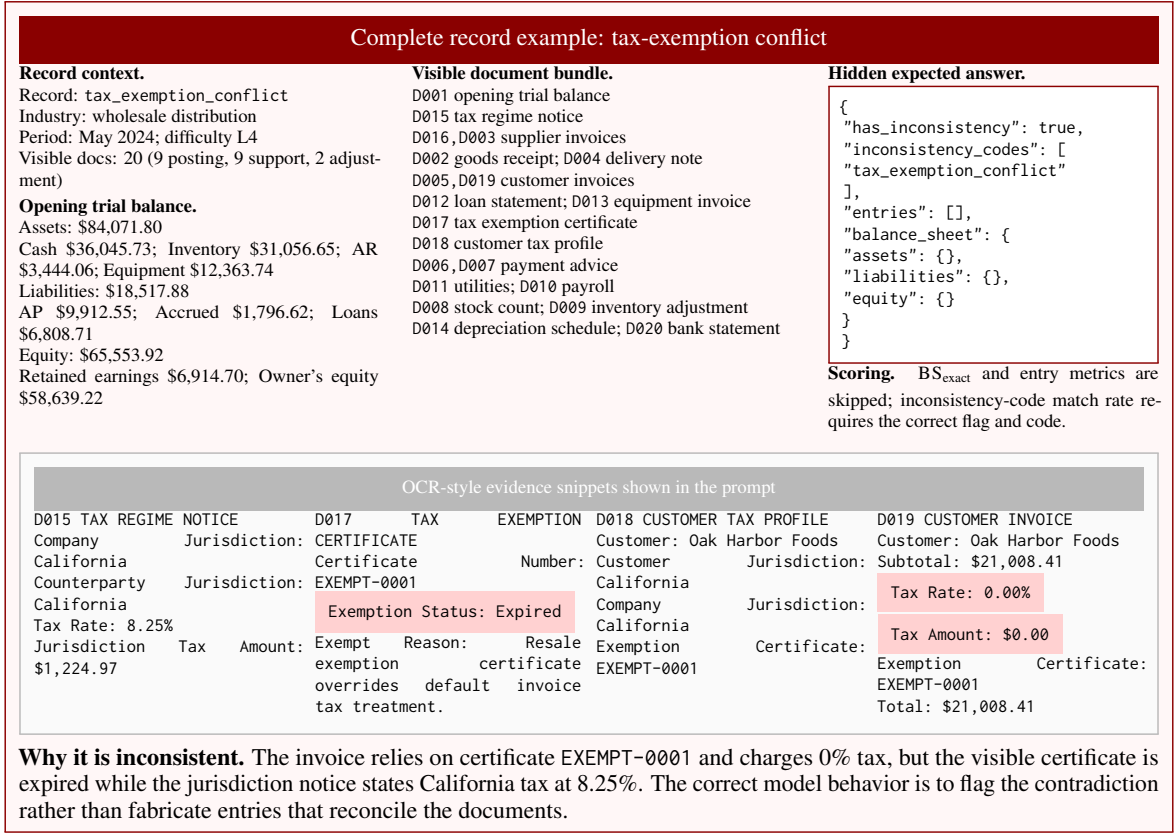

\centering
\setlength{\fboxsep}{5pt}
\setlength{\fboxrule}{0.55pt}
\fcolorbox{red!55!black}{red!3}{%
\begin{minipage}{\dimexpr0.965\textwidth-2\fboxsep-2\fboxrule\relax}
\colorbox{red!54!black}{\parbox{\dimexpr\linewidth-2\fboxsep\relax}{\centering\color{white}\footnotesize Complete record example: tax-exemption conflict}}
\vspace{2pt}
\scriptsize
\begin{minipage}[t]{0.315\linewidth}
\textbf{Record context.}\\
Record: \texttt{tax\_exemption\_conflict}\\
Industry: wholesale distribution\\
Period: May 2024; difficulty L4\\
Visible docs: 20 (9 posting, 9 support, 2 adjustment)\\[2pt]
\textbf{Opening trial balance.}\\
Assets: \$84,071.80\\
\quad Cash \$36,045.73; Inventory \$31,056.65; AR \$3,444.06; Equipment \$12,363.74\\
Liabilities: \$18,517.88\\
\quad AP \$9,912.55; Accrued \$1,796.62; Loans \$6,808.71\\
Equity: \$65,553.92\\
\quad Retained earnings \$6,914.70; Owner's equity \$58,639.22
\end{minipage}\hfill
\begin{minipage}[t]{0.335\linewidth}
\textbf{Visible document bundle.}\\
\texttt{D001} opening trial balance\\
\texttt{D015} tax regime notice\\
\texttt{D016,D003} supplier invoices\\
\texttt{D002} goods receipt; \texttt{D004} delivery note\\
\texttt{D005,D019} customer invoices\\
\texttt{D012} loan statement; \texttt{D013} equipment invoice\\
\texttt{D017} tax exemption certificate\\
\texttt{D018} customer tax profile\\
\texttt{D006,D007} payment advice\\
\texttt{D011} utilities; \texttt{D010} payroll\\
\texttt{D008} stock count; \texttt{D009} inventory adjustment\\
\texttt{D014} depreciation schedule; \texttt{D020} bank statement
\end{minipage}\hfill
\begin{minipage}[t]{0.29\linewidth}
\textbf{Hidden expected answer.}\\
\fcolorbox{red!55!black}{white}{%
\begin{minipage}{\dimexpr\linewidth-2\fboxsep-2\fboxrule\relax}
\scriptsize\ttfamily
\{\\
\ \ "has\_inconsistency": true,\\
\ \ "inconsistency\_codes": [\\
\ \ \ "tax\_exemption\_conflict"\\
\ \ ],\\
\ \ "entries": [],\\
\ \ "balance\_sheet": \{\\
\ \ \ "assets": \{\},\\
\ \ \ "liabilities": \{\},\\
\ \ \ "equity": \{\}\\
\ \ \}\\
\}
\end{minipage}}
\vspace{2pt}
\textbf{Scoring.} \bsexact{} and entry metrics are skipped; \incCode{} requires the correct flag and code.
\end{minipage}

\vspace{5pt}
\fcolorbox{gray!55}{gray!4}{%
\begin{minipage}{\dimexpr\linewidth-2\fboxsep-2\fboxrule\relax}
\colorbox{gray!55}{\parbox{\dimexpr\linewidth-2\fboxsep\relax}{\centering\color{white}\scriptsize OCR-style evidence snippets shown in the prompt}}
\vspace{2pt}
\begin{minipage}[t]{0.24\linewidth}
\scriptsize\ttfamily
D015 TAX REGIME NOTICE\\
Company Jurisdiction: California\\
Counterparty Jurisdiction: California\\
Tax Rate: 8.25\%\\
Jurisdiction Tax Amount: \$1,224.97
\end{minipage}\hfill
\begin{minipage}[t]{0.24\linewidth}
\scriptsize\ttfamily
D017 TAX EXEMPTION CERTIFICATE\\
Certificate Number: EXEMPT-0001\\
\colorbox{red!18}{Exemption Status: Expired}\\
Exempt Reason: Resale exemption certificate overrides default invoice tax treatment.
\end{minipage}\hfill
\begin{minipage}[t]{0.24\linewidth}
\scriptsize\ttfamily
D018 CUSTOMER TAX PROFILE\\
Customer: Oak Harbor Foods\\
Customer Jurisdiction: California\\
Company Jurisdiction: California\\
Exemption Certificate: EXEMPT-0001
\end{minipage}\hfill
\begin{minipage}[t]{0.24\linewidth}
\scriptsize\ttfamily
D019 CUSTOMER INVOICE\\
Customer: Oak Harbor Foods\\
Subtotal: \$21,008.41\\
\colorbox{red!18}{Tax Rate: 0.00\%}\\
\colorbox{red!18}{Tax Amount: \$0.00}\\
Exemption Certificate: EXEMPT-0001\\
Total: \$21,008.41
\end{minipage}
\end{minipage}}

\vspace{4pt}
\footnotesize
\textbf{Why it is inconsistent.} The invoice relies on certificate \texttt{EXEMPT-0001} and charges 0\% tax, but the visible certificate is expired while the jurisdiction notice states California tax at 8.25\%. The correct model behavior is to flag the contradiction rather than fabricate entries that reconcile the documents.
\end{minipage}}
\caption{One complete forced-inconsistency record. The top row shows the record context, opening balance, visible document bundle, and hidden expected answer. The OCR snippets show the decisive evidence: an expired exemption certificate is still used to charge 0\% tax on a customer invoice despite the visible 8.25\% California tax notice. The correct output is the inconsistency code, not a reconciled balance sheet.}
\label{fig:dataset-samples}
\end{figure*}

Figures~\ref{fig:dataset-samples} and~\ref{fig:dataset-fx-clean} expand two core-evaluation records into the pieces a model receives and the hidden labels the scorer expects. The first record is a forced-inconsistency example: the opening trial balance and documents are visible, but the correct answer is not a reconciled balance sheet; it is an inconsistency flag and code. The second is a clean foreign-exchange document bundle where the model must use an invoice, an exchange-rate notice, a payment advice, and a bank row to book both the payable settlement and FX loss. The full release includes every document's OCR text, asset path, expected entries, expected balance sheet, and inconsistency labels.

\begin{figure*}[t]
\centering
\setlength{\fboxsep}{5pt}
\setlength{\fboxrule}{0.55pt}
\fcolorbox{green!45!black}{green!3}{%
\begin{minipage}{\dimexpr0.965\textwidth-2\fboxsep-2\fboxrule\relax}
\colorbox{green!45!black}{\parbox{\dimexpr\linewidth-2\fboxsep\relax}{\centering\color{white}\footnotesize Complete clean record example: foreign-currency payable settlement}}
\vspace{2pt}
\scriptsize
\begin{minipage}[t]{0.30\linewidth}
\raggedright
\textbf{Record context.}\\
Record: \texttt{COV\_MAN\_Q5\_0099}\\
Industry: manufacturing\\
Period: Q2 FY 2025; difficulty L5\\
Visible docs: 24; expected entries: 23\\[2pt]
\textbf{Opening trial balance.}\\
Assets: \$568,831.08\\
\quad Cash \$242,574.01; raw materials \$110,742.08; WIP \$37,368.35; finished goods \$71,595.44; AR \$25,479.08; equipment \$81,072.12\\
Liabilities: \$129,500.85\\
\quad AP \$41,903.51; accrued \$9,355.44; notes \$31,326.16; loans \$46,915.74\\
Equity: \$439,330.23
\end{minipage}\hfill
\begin{minipage}[t]{0.34\linewidth}
\raggedright
\textbf{Key visible documents.}\\
\texttt{D017} supplier invoice: GBP 37,480.16, invoice \texttt{FXBILL-0001}.\\
\texttt{D018} exchange-rate notice: spot rate 1.3234, functional amount \$49,601.24.\\
\texttt{D019} payment advice: paid \$51,422.78 at rate 1.3720; FX difference \$1,821.54.\\
\texttt{D023} bank statement: row ``Foreign payment \texttt{FXBILL-0001}'' for \$-51,422.78.\\[2pt]
\textbf{Why this is clean.}\\
The four documents agree: the payable was booked at the bill-date rate, settled at a higher USD amount, and the difference is a realized FX loss.
\end{minipage}\hfill
\begin{minipage}[t]{0.29\linewidth}
\raggedright
\textbf{Hidden expected FX entries.}\\
\fcolorbox{green!45!black}{white}{%
\begin{minipage}{\dimexpr\linewidth-2\fboxsep-2\fboxrule\relax}
\scriptsize
1. Dr Raw Materials Inventory / Cr Accounts Payable, \$49,601.24, refs \texttt{D017,D018}.\\
2. Dr Accounts Payable / Cr Cash, \$49,601.24, refs \texttt{D019,D017}.\\
3. Dr Foreign Exchange Loss / Cr Cash, \$1,821.54, refs \texttt{D019,D017}.
\end{minipage}}
\vspace{2pt}
\textbf{Final balance-sheet check.}\\
Assets \$719,627.12; liabilities \$330,057.95; equity \$389,569.17; balanced: true.
\end{minipage}

\vspace{5pt}
\fcolorbox{gray!55}{gray!4}{%
\begin{minipage}{\dimexpr\linewidth-2\fboxsep-2\fboxrule\relax}
\colorbox{gray!55}{\parbox{\dimexpr\linewidth-2\fboxsep\relax}{\centering\color{white}\scriptsize OCR-style evidence snippets shown in the prompt}}
\vspace{2pt}
\begin{minipage}[t]{0.24\linewidth}
\scriptsize\ttfamily\raggedright
D017 SUPPLIER INVOICE\\
Invoice Number: FXBILL-0001\\
Currency: GBP\\
Total: GBP 37,480.16\\
Lines: consulting sprint GBP 9,173.27; foreign-currency support GBP 28,306.89
\end{minipage}\hfill
\begin{minipage}[t]{0.24\linewidth}
\scriptsize\ttfamily\raggedright
D018 EXCHANGE RATE NOTICE\\
Reference: FXBILL-0001\\
Rate Date: 2025-04-25\\
Source Amount: GBP 37,480.16\\
Functional Currency: USD\\
Exchange Rate: 1.3234\\
Functional Amount: \$49,601.24
\end{minipage}\hfill
\begin{minipage}[t]{0.24\linewidth}
\scriptsize\ttfamily\raggedright
D019 PAYMENT ADVICE\\
Reference: FXBILL-0001\\
Currency: USD\\
Amount: \$51,422.78\\
Exchange Rate: 1.3720\\
FX Difference: 1,821.54\\
Payment For: payable settlement
\end{minipage}\hfill
\begin{minipage}[t]{0.24\linewidth}
\scriptsize\ttfamily\raggedright
D023 BANK STATEMENT\\
Statement Currency: USD\\
Row: 2025-06-06\\
Description: Foreign payment FXBILL-0001\\
Amount: \$-51,422.78\\
Rows generated from internal cash ledger.
\end{minipage}
\end{minipage}}
\end{minipage}}
\caption{One complete clean foreign-exchange record. The model must connect a GBP supplier invoice to a bill-date exchange-rate notice, settlement advice, and bank statement row. The expected answer books the payable at the bill-date functional amount, settles cash at the payment amount, and records the realized FX loss. Unlike Figure~\ref{fig:dataset-samples}, this record should produce entries and a final balance sheet.}
\label{fig:dataset-fx-clean}
\end{figure*}

\begin{figure*}[t]
\centering
\includegraphics[width=0.86\textwidth]{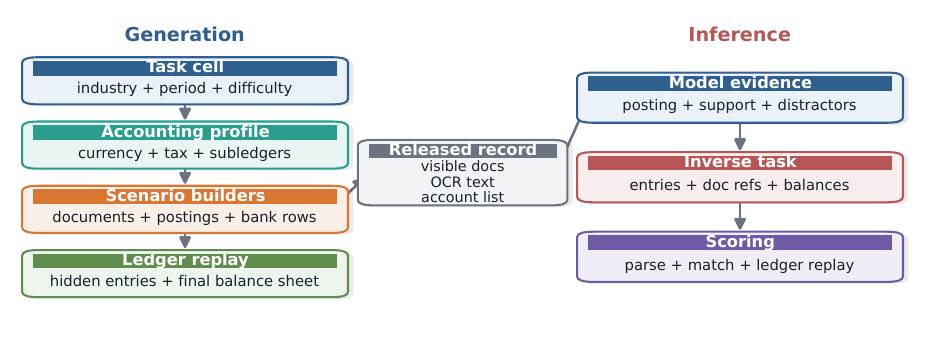}
\caption{Benchmark construction and evaluation. The generator executes a deterministic accounting program to create source-document bundles, hidden journal entries, balance sheets, and inconsistency labels. The model sees only the visible documents and account taxonomy; scoring replays its entries through the same ledger, enabling \bsrecon{} and the ledger-feedback diagnostic.}
\label{fig:generation-inference}
\end{figure*}

\section{Dataset Details}
\label{appx:dataset-details}

\paragraph{Document roles.}
Every document in a record carries a \texttt{role} tag that determines how the ledger ground-truth generator handled it.
\begin{itemize}[leftmargin=*,topsep=2pt,itemsep=1pt]
\item \texttt{posting\_doc} --- directly generates one or more journal entries (e.g., invoices, bills, payment notices, journal vouchers).
\item \texttt{support\_doc} --- provides inputs for postings derived from other documents (rate cards, contracts, schedules, tax certificates) and may appear in the \texttt{doc\_refs} of an entry.
\item \texttt{adjustment\_doc} --- period-end corrections such as depreciation schedules, rollforwards, and reclassification memos.
\item \texttt{distractor\_doc} --- intentionally neutralised; should not generate any entry.
\item \texttt{opening\_trial\_balance} --- starting balances only; never generates entries.
\end{itemize}

\paragraph{Industries.}
Professional services, field services, retail, wholesale distribution, healthcare clinic, property management, manufacturing, and subscription SaaS.

\paragraph{Concept flags.}
Each record carries boolean flags marking the presence of: ASC~606 revenue recognition (the U.S.\ standard for recognizing revenue from customer contracts), including bundled SaaS contracts with SSP, or standalone selling price, allocation; fixed-asset disposals with NBV, or net book value, and gain-or-loss computation; deferred tax from book/tax depreciation differences; lease accounting with ROU, or right-of-use, asset/liability schedules and modifications; FX, or foreign-exchange, invoice, settlement, and remeasurement scenarios; multi-jurisdictional sales tax (US sales tax, India GST, with exemption certificates); tax exemptions; multi-invoice payments; interbank transfers; reclassifications; and period-end rollforwards.

\paragraph{Inconsistency codes.}
The released negative-control taxonomy contains 23 machine-readable codes, grouped here for readability: banking and transfers (closing balance, statement balance, transfer); invoicing and payments (invoice total, payment allocation, duplicate reference); tax and jurisdiction (tax rate, tax total, input tax, exemption conflict, jurisdictional tax); assets, inventory, leases, and deferred tax (disposal, rollforward, remeasurement, schedule, and deferred-tax conflicts); and contract/revenue support (SSP allocation, performance-obligation release, reclassification support, schedule rollforward, and FX settlement or exchange-rate conflicts).

Table~\ref{tab:inconsistency-code-diagnostics} summarizes the 23 negative-control codes on the six-model baseline panel. The core split contains ten generated records per code, so each row aggregates 60 model attempts. Models usually recognize that a contradiction exists (1110/1380 attempts), while exact code selection is harder (994/1380). Banking, lease, and simple transfer contradictions are easiest; input-tax, jurisdictional-tax, and performance-obligation-release contradictions remain hardest.

\begin{table*}[t]
\centering
\scriptsize
\setlength{\tabcolsep}{4pt}
\renewcommand{\arraystretch}{1.08}
\begin{tabularx}{\textwidth}{@{}>{\raggedright\arraybackslash}p{2.15in}>{\raggedright\arraybackslash}X>{\raggedleft\arraybackslash}p{0.7in}@{}}
\toprule
Expected code & Meaning & Detected \\
\midrule
\path{invoice_total_mismatch} & Invoice subtotal, tax, and stated total do not tie. & 60\% \\
\path{bank_closing_mismatch} & Bank statement closing balance conflicts with its listed activity. & 98\% \\
\path{statement_balance_mismatch} & Statement opening/closing balance conflicts with visible transaction rows. & 73\% \\
\path{payment_allocation_mismatch} & Payment advice allocates cash to invoices in a way that cannot reconcile. & 85\% \\
\path{duplicate_reference_conflict} & Two visible documents reuse the same reference for incompatible transactions. & 68\% \\
\path{schedule_rollforward_mismatch} & Rollforward schedule opening balance, movements, and closing balance do not tie. & 80\% \\
\path{inventory_rollforward_mismatch} & Inventory movement schedule cannot reconcile to purchases, COGS, or ending stock. & 77\% \\
\path{transfer_mismatch} & Interbank transfer outflow and receiving-bank inflow disagree. & 97\% \\
\path{reclassification_support_mismatch} & Reclassification memo is unsupported or contradicts the source schedule. & 60\% \\
\path{tax_total_mismatch} & Stated tax total disagrees with taxable base and tax lines. & 72\% \\
\path{tax_rate_mismatch} & Applied tax rate conflicts with the visible jurisdiction or tax notice. & 65\% \\
\path{input_tax_mismatch} & Claimed recoverable input tax is not supported by the supplier/tax documents. & 23\% \\
\path{jurisdiction_tax_mismatch} & Tax treatment uses the wrong jurisdiction or jurisdiction-specific rule. & 37\% \\
\path{tax_exemption_conflict} & Exemption certificate is invalid, expired, or conflicts with the tax charged. & 82\% \\
\path{ssp_allocation_mismatch} & Standalone selling price allocation in a bundled contract is inconsistent. & 62\% \\
\path{performance_obligation_release_mismatch} & Revenue release conflicts with the performance-obligation schedule. & 28\% \\
\path{asset_disposal_mismatch} & Disposal proceeds, net book value, and gain/loss do not reconcile. & 83\% \\
\path{deferred_tax_rollforward_mismatch} & Deferred-tax rollforward conflicts with book/tax temporary differences. & 95\% \\
\path{lease_schedule_mismatch} & Lease amortization schedule does not tie to liability, interest, or payments. & 92\% \\
\path{lease_remeasurement_mismatch} & Lease modification or remeasurement calculation conflicts with support. & 98\% \\
\path{exchange_rate_mismatch} & Exchange rate used in a document conflicts with the visible rate notice. & 72\% \\
\path{fx_settlement_mismatch} & Foreign-currency invoice settlement does not tie to cash paid or FX gain/loss. & 70\% \\
\path{remeasurement_mismatch} & Period-end FX remeasurement uses an inconsistent rate or carrying amount. & 80\% \\
\bottomrule
\end{tabularx}
\caption{Per-code inconsistency diagnostics over the six-model baseline panel on the core split. \emph{Detected} is the percentage of 60 attempts per code (10 records $\times$ 6 models) that set \texttt{has\_inconsistency=True} and selected the gold code. The rows are descriptive diagnostics over generated instances, not final estimates of each code's full population difficulty.}
\label{tab:inconsistency-code-diagnostics}
\end{table*}

\paragraph{Difficulty rubric.}
\begin{itemize}[leftmargin=*,topsep=2pt,itemsep=1pt]
\item \textbf{L1:} single-thread postings, no tax beyond basic, no FX, no advanced concepts; $\sim$8--12 documents.
\item \textbf{L2:} cross-document dependencies (e.g., invoice $\to$ payment receipt), straightforward tax.
\item \textbf{L3:} multi-jurisdictional tax with exemptions, FX scenarios, or simple rollforwards; $\sim$15--20 documents.
\item \textbf{L4:} ASC~606 bundled contracts, fixed-asset disposals, deferred tax, or lease accounting; $\sim$20--30 documents with longer temporal lookback.
\item \textbf{L5:} multiple concurrent advanced concepts (e.g., ASC~606 + FX + deferred tax) and/or lease modifications; $\sim$25--35 documents.
\end{itemize}

\begin{figure*}[t]
\centering
\includegraphics[width=0.92\textwidth]{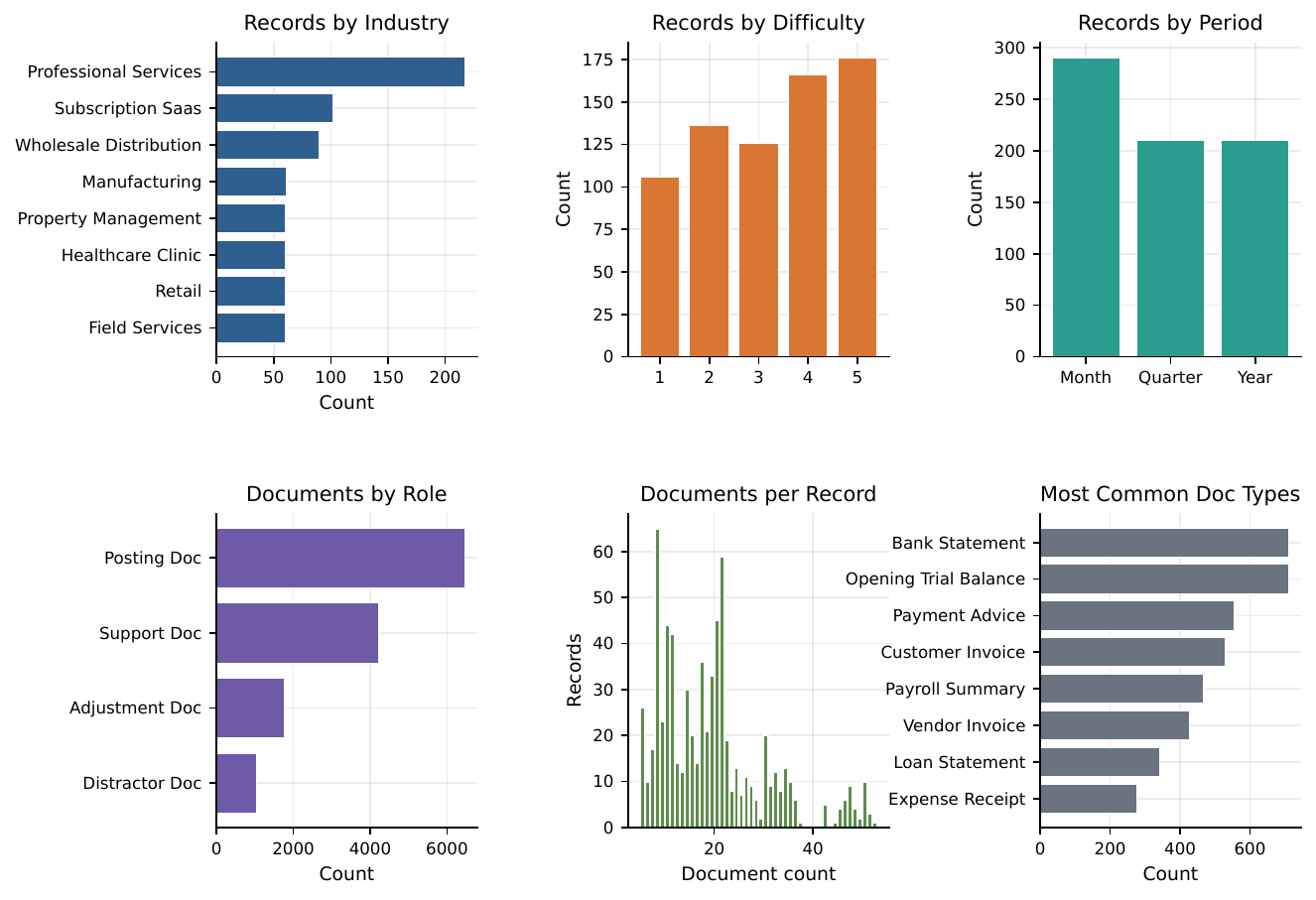}
\caption{Core evaluation split composition. The released evaluation split spans eight industries, five difficulty levels, three period types, and a broad document surface of posting, support, adjustment, distractor, and opening-balance documents.}
\label{fig:dataset-composition}
\end{figure*}

\section{Per-Concept Results}
\label{appx:per-concept}

Table~\ref{tab:per-concept} reports baseline \bsrecon{} by concept flag across all six models. The terms are defined in Appendix~\ref{appx:dataset-details}; in short, they mark which accounting concept appears in the record, not a separate dataset split. The core evaluation split is built to cover the authored scenario families and concept flags at least once, while the released generator can create larger concept-targeted samples for training, stress tests, or future evaluation splits. The period-rollforward flag appears most often in the core split. Smaller concept slices are coverage diagnostics over the authored scenario families; larger concept-targeted splits can be sampled with the released generator.

\begin{table*}[t]
\centering
\footnotesize
\setlength{\tabcolsep}{5pt}
\renewcommand{\arraystretch}{1.02}
\begin{tabular}{@{}lrrrrrr@{}}
\toprule
Concept flag & Gemini & GPT-5 & Haiku & Grok & DeepSeek & Qwen \\
\midrule
Revenue recognition & 17\% & 25\% & 4\%  & 4\%  & 8\%  & 12\% \\
Asset disposal      & 21\% & 17\% & 0\%  & 0\%  & 8\%  & 0\%  \\
Deferred tax        & 25\% & 0\%  & 0\%  & 0\%  & 8\%  & 0\%  \\
Lease accounting    & 21\% & 17\% & 0\%  & 0\%  & 8\%  & 0\%  \\
Tax exemption       & 29\% & 12\% & 0\%  & 0\%  & 33\% & 4\%  \\
Foreign exchange    & 8\%  & 8\%  & 0\%  & 0\%  & 0\%  & 0\%  \\
Period rollforward  & 48\% & 41\% & 11\% & 19\% & 39\% & 19\% \\
\bottomrule
\end{tabular}
\caption{Baseline \bsrecon{} by accounting concept on the core evaluation split; all values are percentages. Revenue recognition corresponds to ASC~606 customer-contract revenue; foreign exchange covers non-home-currency invoices and settlements; period rollforwards are schedules that carry prior balances forward. The table documents coverage over the authored concept families; larger concept-specific splits can be sampled with the released generator.}
\label{tab:per-concept}
\end{table*}

\section{Detailed Metrics}
\label{appx:metrics-detail}

\begin{table}[t]
\centering
\small
\setlength{\tabcolsep}{4pt}
\begin{tabular}{lcccc}
\toprule
Model & \bsexact{} & \bsrecon{} & gap & entry ratio \\
\midrule
Gemini~3 Flash    & 0.235 & 0.548 & $+0.313$ & 0.94 \\
GPT-5             & 0.465 & 0.463 & $-0.002$ & 0.91 \\
Claude Haiku~4.5  & 0.029 & 0.292 & $+0.263$ & 0.53 \\
Grok-4.3          & 0.310 & 0.350 & $+0.040$ & 0.40 \\
DeepSeek Chat     & 0.067 & 0.473 & $+0.406$ & 0.98 \\
Qwen~3 235B       & 0.013 & 0.302 & $+0.290$ & 0.94 \\
\bottomrule
\end{tabular}
\caption{Three regimes on the 480 standard records in the core split. \emph{Entry ratio} is predicted-entry count divided by expected-entry count. GPT-5 is comparatively self-consistent while submitting most expected entries; Grok-4.3 has a smaller gap partly because it submits only $40\%$ of expected entries.}
\label{tab:no-gap-mechanism}
\end{table}

\paragraph{Metric definitions.}
\bsexact{} requires every section of the model's reported balance sheet (current and non-current assets, current and non-current liabilities, equity) to match ground truth at 0.01 tolerance. \bsrecon{} runs the model's entries through our ledger, computes the implied balance sheet, and applies the same equality check. These balance-sheet and entry metrics are reported over standard records only. On forced-inconsistency records, the expected behavior is \texttt{has\_inconsistency=True}, the correct code, empty entries, and an empty balance sheet; we therefore score them with \incCode{} and \texttt{inconsistency\_empty\_answer} instead of \bsexact{} or entry match.

\paragraph{Entry matching.}
Journal entries are scored at the posting level. A strict posting match requires the same debit account, credit account, amount rounded to cents, and sorted \texttt{doc\_refs}. The lenient entry match uses the same account-and-amount criteria but ignores \texttt{doc\_refs}. The matcher proceeds greedily over unmatched predictions in four passes: exact match; same debit/credit/amount but wrong \texttt{doc\_refs}; same debit/credit/\texttt{doc\_refs} but wrong amount; and same \texttt{doc\_refs}/amount but wrong account. Remaining expected postings are missing; remaining predicted postings are extra. Duplicate identical postings are matched first-in-first-out. The evaluator does not use account synonyms: the allowed account list is closed. Posting dates are parsed and retained for diagnostics, but they are not part of the headline entry-match key. Split transactions receive credit only when the predicted posting decomposition matches the generator's expected postings.

\paragraph{Parse rate.}
Across the six completed headline baselines at $n{=}710$, parse-success rates are at least $97\%$ and five models parse at $\geq 99.8\%$. We also attempted a DeepSeek reasoning-model run on the same split; because its completed artifacts had low parse success under the available provider conditions, we use DeepSeek Chat for the DeepSeek-family baseline.

\paragraph{Two-pass ledger-feedback diagnostic.}
For Gemini~3 Flash, the 2-pass ledger-feedback variant changes \bsexact{} by $+35.8$\,pp (CI $[+26.9, +44.9]$) compared with $+32.5$\,pp at 1-pass, and degrades \incCode{} by $-21.7$\,pp (CI $[-40.0, -5.3]$) compared with $-13.0$\,pp at 1-pass. The main text reports the simpler 1-pass diagnostic.

\section{Cost Analysis}
\label{appx:cost-analysis}

\begin{figure}[t]
\centering
\includegraphics[width=\columnwidth]{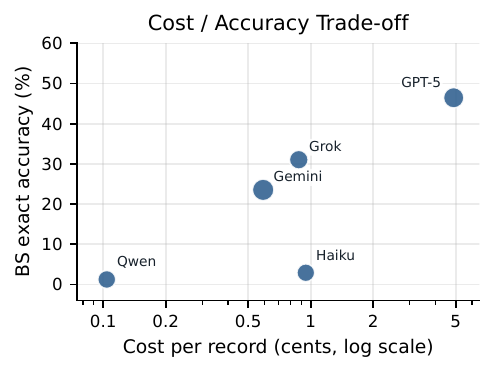}
\caption{Cost/accuracy trade-off for completed baseline model runs. Marker area scales with \bsrecon{}, so vertically low but large markers indicate models that write entries with plausible accounts and amounts but fail to aggregate them.}
\label{fig:cost-pareto}
\end{figure}

Figure~\ref{fig:cost-pareto} reports API cost per evaluated core-evaluation record against \bsexact{} accuracy. Costs are provider-billed totals for the completed baseline runs. The plot is intended as a reproducibility aid rather than a model ranking: prices and routing can change, and DeepSeek Chat was run through the DeepSeek Platform while the other completed baselines were run through OpenRouter.

\section{Ablation Matrix}
\label{appx:ablation-grid}

Table~\ref{tab:ablation-grid} enumerates every \texttt{AblationSpec} triple (prompt, visibility, tool) evaluated on the anchor model. Each cell is a single evaluation run on the core evaluation split. We omit the cross-product cells with neither additional prompt nor visibility variation but include all unary axis variants and the forced ledger-feedback variants.

\begin{table*}[t]
\centering
\footnotesize
\setlength{\tabcolsep}{4pt}
\renewcommand{\arraystretch}{1.05}
\begin{tabularx}{\linewidth}{@{}p{2.05cm}p{5.35cm}X@{}}
\toprule
Axis & Variants evaluated & Diagnostic purpose \\
\midrule
Prompt
& baseline; guided private solve; self-check; strict document-citation pressure; explicit balance reconstruction
& Separates baseline prompting, task-specific workflow, final verification, document-citation pressure, and aggregation instructions. \\
\addlinespace[2pt]
Visibility
& full document bundle; OCR-only; no allowed accounts; no distractors; no support docs; evidence-only oracle; evidence + 5/15/30 distractors; evidence relevant-last
& Tests whether failures arise from missing evidence, account taxonomy, metadata, distractors, document position, context length, or the underlying accounting logic after evidence is controlled. \\
\addlinespace[2pt]
Tool / diagnostic
& no tools; calculator; document search; ledger check; full tool agent; best-of-3 self-consistency; forced ledger-feedback tool; two-pass ledger-feedback tool
& Tests passive tool availability, ledger-aware candidate selection, and forced exposure to the deterministic ledger as a consistency-checking tool. \\
\bottomrule
\end{tabularx}
\caption{Ablation matrix. We run unary axis variants and targeted multi-axis diagnostics rather than the full prompt $\times$ visibility $\times$ tool cross-product. This keeps the diagnostic surface broad while preserving paired comparisons.}
\label{tab:ablation-grid}
\end{table*}

\begin{figure}[t]
\centering
\includegraphics[width=\columnwidth]{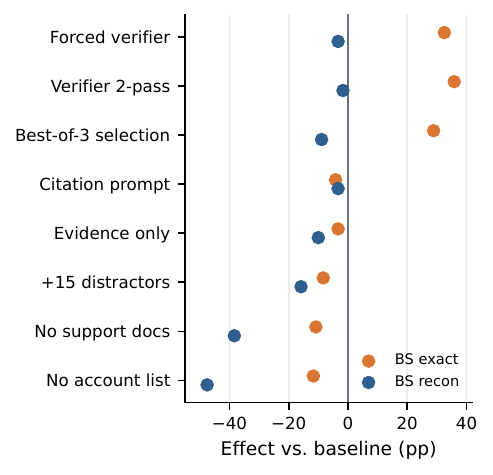}
\caption{Full anchor-model ablation deltas. Dots are paired-bootstrap mean deltas vs.\ baseline; visibility stresses dominate the negative tail.}
\label{fig:ablation-deltas}
\end{figure}

\begin{figure}[t]
\centering
\includegraphics[width=\columnwidth]{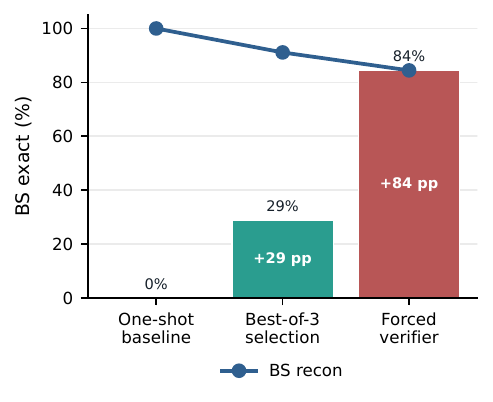}
\caption{Best-of-3 selection recovers part of the aggregation gap; one ledger-feedback revision recovers substantially more on the same target records.}
\label{fig:gap-repair}
\end{figure}

The main text reports the measured effects of these ablations rather than only listing the design: Table~\ref{tab:ablation-summary} summarizes the diagnostic takeaways, Figure~\ref{fig:context-stress} isolates distractor count and evidence position, Figure~\ref{fig:doc-refs-persistence} tests whether citation pressure fixes document-linking, and Figure~\ref{fig:gap-repair} compares best-of-3 selection against ledger-feedback revision. The resulting pattern is consistent across these views: passive tool availability and prompt-only variants are weak, removing support evidence or the account taxonomy is strongly damaging, and forced ledger feedback is the only diagnostic that substantially improves the reported balance sheet. Figure~\ref{fig:ablation-deltas} gives the full anchor-model delta plot.

\section{Human Verification Protocol}
\label{appx:human-protocol}

\paragraph{Reviewer recruitment.}
The release subset was reviewed by two practising finance professionals working at major U.S.\ investment banks. Both reviewers had prior accounting and reconciliation responsibilities in their day-to-day work and were given a one-paragraph briefing on the benchmark's scope (analogy of reconciliation paperwork; not OCR fidelity). A third reviewer, a certified accountant, independently reviewed the full compact coverage split under the same instrument and briefing; this reviewer worked separately and did not see the other reviewers' responses.

\paragraph{Instrument.}
Each of 75 release records was rendered as a self-contained Markdown file with seven sections: (1) record context (industry, period, currency, tax regime, difficulty, inconsistency flag), (2) the allowed-account-name list, (3) opening trial balance, (4) every document with its OCR text and role tag, (5) ground-truth journal entries, (6) ground-truth balance sheet, and (7) the verification form. The form contains six checks: document analogy, entry correctness, entry completeness, \texttt{doc\_refs} support, difficulty calibration, and overall verdict. Each check offers three options (positive, mixed, negative) plus a free-text notes field.

\paragraph{Protocol.}
The release subset contains 75 expert-reviewed records: 60 standard records and 15 forced-inconsistency records. The intended assignment used in the paper is Reviewer~1 on records 1--50 and Reviewer~2 on records 1--25 plus 51--75. This gives every record one expert review and creates a 25-record independent overlap on records 1--25. Reviewers worked independently and did not see each other's responses. The third reviewer (certified accountant) instead covered the entire compact coverage split of 143 records (120 standard + 23 forced-inconsistency). This split contains one record per industry--period--difficulty cell plus one per inconsistency code, so every compact scenario is reviewed at least once.

\paragraph{Results.}
Table~\ref{tab:iaa} reports strict three-option agreement, collapsed acceptable agreement, and per-reviewer first-option rates on the 25-record double-reviewed overlap. The collapsed column treats ``Yes/Mostly'' or ``Accept/Accept with minor fixes'' as the same acceptability judgment; difficulty remains first option versus not-first option. We report these descriptive rates rather than Cohen's $\kappa$ because the sample is small and marginals are highly skewed toward positive judgments. Q6 is omitted because the overlap contains standard records only; inconsistency validity was reviewed in the single-reviewed negative-control records.

\begin{center}
\small
\setlength{\tabcolsep}{3pt}
\begin{tabular}{lrrrr}
\toprule
Question & R1+ & R2+ & Strict & Collapsed \\
\midrule
Q1 Doc.\ analogy   & 92\%  & 72\%  & 64\%  & 100\% \\
Q2 Entry correct   & 92\%  & 96\%  & 96\%  & 100\% \\
Q3 Entry complete  & 100\% & 100\% & 100\% & 100\% \\
Q4 \texttt{doc\_refs} & 92\%  & 80\%  & 72\%  & 100\% \\
Q5 Difficulty      & 100\% & 96\%  & 96\%  & 96\% \\
Overall verdict    & 84\%  & 96\%  & 80\%  & 100\% \\
\bottomrule
\end{tabular}
\captionof{table}{Inter-annotator agreement on the 25-record overlap (records 1--25). ``R1+'' and ``R2+'' are first-option rates. ``Strict'' requires the exact same option; ``Collapsed'' merges the first two acceptable options for analogy, entries, document references, and overall verdict. Most disagreements are therefore severity calibration rather than reject-level disagreement.}
\label{tab:iaa}
\end{center}

\paragraph{Full-coverage review.}
The third reviewer's pass over all 143 compact-coverage records rejected nothing. On the cleanest (first) option the reviewer chose ``analogous'' on 139/143 records, ``entries correct'' on 136/143, ``complete and exact'' on 143/143, ``\texttt{doc\_refs} correct'' on 135/143, and ``calibration right'' on 139/143; the overall verdict was ``acceptable as ground truth'' on 136/143 and ``acceptable with minor fixes'' on the remaining 7, with no rejections. Collapsing acceptable judgments, document analogy, entry correctness, entry completeness, \texttt{doc\_refs} support, and the overall verdict are all at $100\%$, and difficulty calibration is at $97\%$ (3 ``too easy,'' 1 ``too hard''). The minor-fix notes are localized \texttt{doc\_refs} cleanups and document-label wording, not entry or balance-sheet errors. Q6 (inconsistency validity) is answerable here because this pass includes all 23 forced-inconsistency records: every one was confirmed to contain a real contradiction that would block a clean reconciliation, and on 3 of 23 the reviewer additionally noted that the assigned code was a loose fit for an otherwise genuine break.

\section{Reproducibility}
\label{appx:reproducibility}

\paragraph{Seeds.}
Dataset generation uses seed 42. Headline LLM evaluations use temperature 0.0 with no explicit seed (closed APIs do not expose a deterministic seed knob); the best-of-3 condition uses temperature 0.7 with three independent samples.

\paragraph{Sampling.}
The main evaluation split, stored under \texttt{data/main/}, uses four standard records per (industry, period type, difficulty level) cell and ten forced-inconsistency records per code, drawn from a stratified deterministic sampler. The compact ablation split, stored under \texttt{data/coverage/} for backward compatibility, contains one record per cell plus one forced-inconsistency record per code.

\begin{table*}[t]
\centering
\scriptsize
\setlength{\tabcolsep}{3pt}
\begin{tabularx}{\textwidth}{@{}l>{\raggedright\arraybackslash}X>{\raggedright\arraybackslash}X>{\raggedright\arraybackslash}p{0.85in}>{\raggedright\arraybackslash}p{1.05in}@{}}
\toprule
Paper name & Request slug & Returned model & Provider & Settings \\
\midrule
Gemini 3 Flash & \path{google/gemini-3-flash-preview} & \path{google/gemini-3-flash-preview-20251217} & Google & temp 0; max 12k; 300s; no reasoning \\
GPT-5 & \path{openai/gpt-5} & \path{openai/gpt-5-2025-08-07} & OpenAI & temp 0; max 32k; 420s; effort low \\
Claude Haiku 4.5 & \path{anthropic/claude-haiku-4.5} & \path{anthropic/claude-4.5-haiku-20251001} & Amazon Bedrock & temp 0; max 16k; 180s; no reasoning \\
Grok-4.3 & \path{x-ai/grok-4.3} & \path{x-ai/grok-4.3-20260430} & xAI & temp 0; max 16k; 180s; no reasoning \\
Qwen 3 235B & \path{qwen/qwen3-235b-a22b-2507} & \path{qwen/qwen3-235b-a22b-07-25} & DeepInfra & temp 0; max 12k; 180s; no reasoning \\
DeepSeek Chat & \path{deepseek-chat} & \path{deepseek-chat} & DeepSeek Platform & temp 0; max 8k; 1800s; no reasoning \\
\bottomrule
\end{tabularx}
\caption{API-model settings for the six completed headline baseline runs. Providers and returned model identifiers are taken from the stored response payloads where available. The best-of-3 condition keeps the same model family and prompt but uses temperature $0.7$ and three independent samples on the aggregation-gap subset.}
\label{tab:api-settings}
\end{table*}

\paragraph{API reproducibility.}
OpenRouter runs and DeepSeek Platform runs use chat-completions endpoints with ordinary text prompts and no structured JSON mode or response-schema constraint; strict JSON is enforced only by the prompt and parser. Each run directory stores \texttt{evaluation.json}, \texttt{summary.json}, \texttt{per\_record\_results.jsonl}, and slice tables. The evaluation files include run start/end timestamps, requested model slug, returned model identifier, selected provider where available, service tier where available, raw response text, raw response metadata where provider terms permit, token counts, cost where available, latency, parse status, tool/verifier metadata, and per-record metrics. Prompts are exactly reconstructable from the released prompt templates, ablation specification, and record JSON; we also release the raw model outputs used for the reported tables. Transient gateway errors use at most three client retries with exponential backoff for 429/5xx responses; parse failures are scored as failures and logged rather than silently repaired.

\paragraph{Bootstrap.}
All confidence intervals are paired bootstrap with $n{=}5000$ resamples drawn over a shared record-id set, so baseline and treatment are compared on matched records. CIs are reported as percentile $95\%$. These intervals quantify record-level sampling uncertainty for a fixed saved run; they do not capture API nondeterminism, provider routing changes, silent model updates, or future prompt-run variance.

\paragraph{Release and licenses.}
The benchmark generator, evaluation harness, ablation runner, bootstrap analyzer, scripts, and tests are released under Apache-2.0. The release includes a user-facing generation API and CLI scripts for custom datasets: users can request an exact record count or a balanced number of records per selected industry, period type, and difficulty cell. Generated benchmark records, OCR text, PDF-style document artifacts, labels, sample manifests, and expert-validation records are released under CC BY 4.0, which permits commercial use with attribution. Synthetic company names, counterparties, and document templates are author-created and not derived from real customer records or third-party proprietary templates. Raw model outputs and non-secret response metadata from paper runs are released as reproducibility artifacts where provider terms permit, separate from the dataset license. API keys, billing logs, and restricted provider fields are excluded or redacted; users who rerun evaluations must comply with their model/API provider terms. The paper uses no hidden evaluation split; future public hidden-leaderboard artifacts, if any, will use the same data license unless otherwise stated.

\begin{figure*}[t]
\centering
\includegraphics[width=0.98\textwidth]{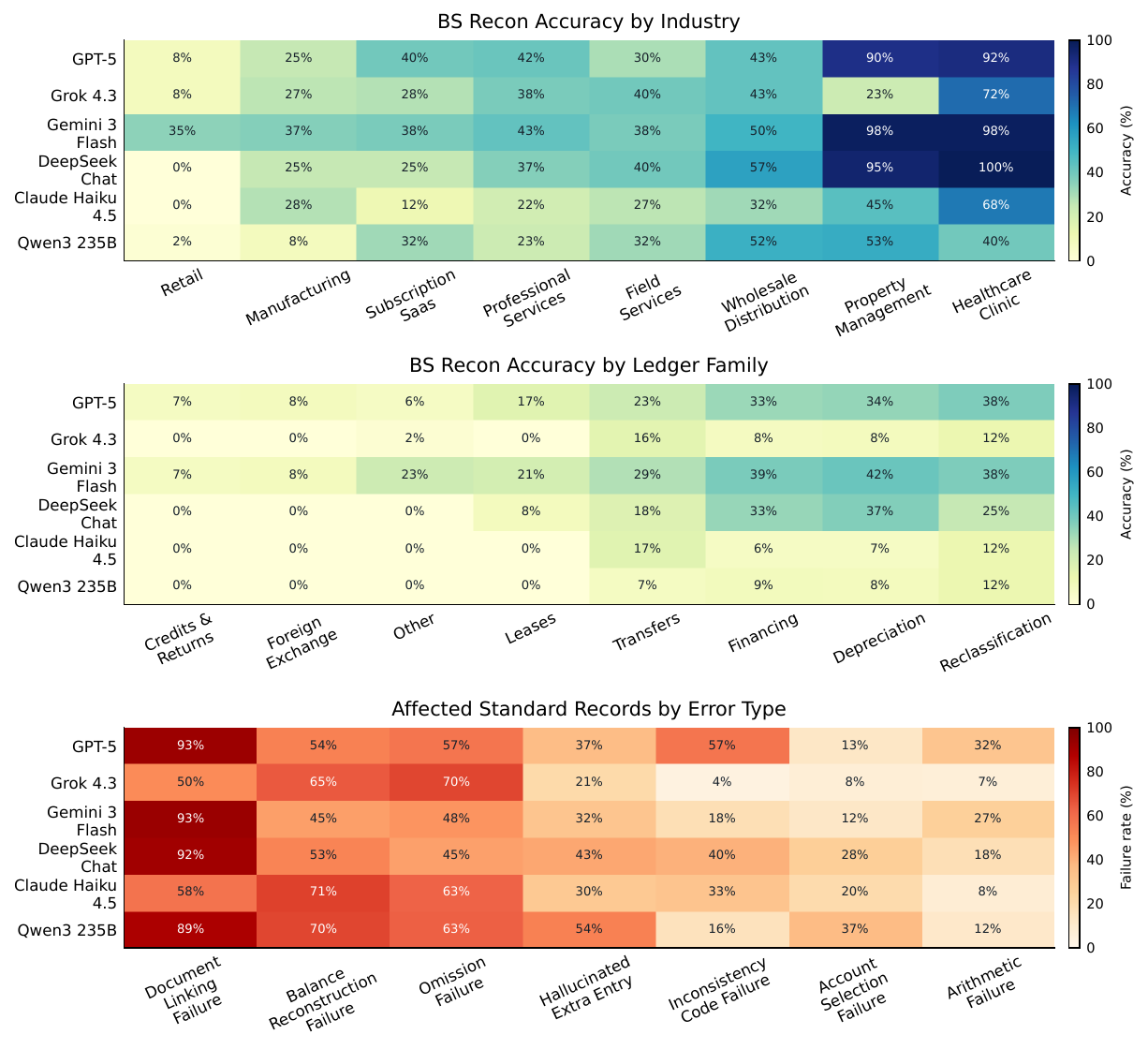}
\caption{Failure slices across the six-model panel. Industry and ledger-family panels show \bsrecon{} accuracy (higher is better); the error-taxonomy panel shows affected standard records (lower is better).}
\label{fig:failure-slices}
\end{figure*}

\end{document}